\documentclass[11pt]{article}

\usepackage[final]{acl}

\usepackage{times}
\usepackage{latexsym}

\usepackage[T1]{fontenc}

\usepackage[utf8]{inputenc}

\usepackage{microtype}

\usepackage{inconsolata}
\usepackage{kotex}
\usepackage[textsize=small]{todonotes}
\usepackage{graphicx}
\usepackage{multirow}
\usepackage[table,xcdraw]{xcolor}
\usepackage{tabularx}
\usepackage{amssymb}
\usepackage{pifont}
\usepackage{booktabs}
\usepackage{hyperref}
\usepackage[most]{tcolorbox}
\usepackage{xcolor}
\usepackage{algorithm}
\usepackage{algpseudocode}
\usepackage{enumitem}
\usepackage{soul}
\usepackage{makecell}

\usepackage[dvipsnames]{xcolor}

\definecolor{navy}{RGB}{0, 75, 150}
\definecolor{coral}{RGB}{210, 25, 20}

\definecolor{myblue}{RGB}{99,240,255}

\definecolor{blue1}{RGB}{235,245,255}   
\definecolor{blue2}{RGB}{210,230,250} 
\definecolor{blue3}{RGB}{180,210,240} 
\definecolor{blue4}{RGB}{140,180,225} 
\definecolor{blue5}{RGB}{155,165,255}    
\definecolor{blue6}{RGB}{135,145,255}    
\definecolor{blue7}{RGB}{115,125,255}    %

\definecolor{red1}{RGB}{255,235,235}   
\definecolor{red2}{RGB}{255,210,210}   
\definecolor{red3}{RGB}{255,175,175}   

\newcommand{\cmark}{\textcolor{green!60!black}{\checkmark}}
\newcommand{\xmark}{\textcolor{red!70!black}{\ding{55}}}

\newtcolorbox{promptbox}{
    enhanced,
    colback=white,
    colframe=black,
    coltitle=white,
    boxrule=0.5pt,
    rounded corners,
    sharp corners=downhill,   
    attach boxed title to top center={
        yshift=-3mm           
    },
    boxed title style={
        colback=black,
        colframe=black,
        top=2mm, bottom=2mm,
        left=3mm, right=3mm,
        font=\bfseries\large,
        width=\linewidth      
    },
    top=5mm, bottom=5mm, left=5mm, right=5mm,
}

\title{FeedEval: Pedagogically Aligned Evaluation of LLM-Generated Essay Feedback}

\author{
  Seongyeub Chu\textsuperscript{1}, 
  Jongwoo Kim\textsuperscript{2},  
  Mun Yong Yi\textsuperscript{1}\thanks{Corresponding author.} \\
  \textsuperscript{1}Graduate School of Data Science, KAIST \\
  \textsuperscript{2}Department of Industrial \& Systems Engineering, KAIST \\
  \texttt{\{chseye7, gsds4885, munyi\}@kaist.ac.kr}
}

\begin{document}
\maketitle
\begin{abstract}
Going beyond the prediction of numerical scores, recent research in automated essay scoring has increasingly emphasized the generation of high-quality feedback that provides justification and actionable guidance. To mitigate the high cost of expert annotation, prior work has commonly relied on LLM-generated feedback to train essay assessment models. However, such feedback is often incorporated without explicit quality validation, resulting in the propagation of noise in downstream applications. To address this limitation, we propose FeedEval, an LLM-based framework for evaluating LLM-generated essay feedback along three pedagogically grounded dimensions: specificity, helpfulness, and validity. FeedEval employs dimension-specialized LLM evaluators trained on datasets curated in this study to assess multiple feedback candidates and select high-quality feedback for downstream use. Experiments on the ASAP++ benchmark show that FeedEval closely aligns with human expert judgments and that essay scoring models trained with FeedEval-filtered high-quality feedback achieve superior scoring performance. Furthermore, revision experiments using small LLMs show that the high-quality feedback identified by FeedEval leads to more effective essay revisions. We release our code and curated datasets at: \textbf{\url{https://github.com/BBeeChu/FeedEval.git}}.
\end{abstract}

\section{Introduction}
\label{intro}
Automated essay assessment has evolved from feature-engineered approaches to pre-trained and large language models (LLMs) \cite{ramesh2022automated, li2024automated, misgna2024survey}. While early work focused primarily on essay scoring, recent studies have explored joint modeling of scoring and feedback generation to provide pedagogically meaningful guidance beyond numerical scores. Developing essay feedback generation models typically requires pedagogically grounded feedback annotated by domain experts, which is costly and impractical in real-world settings \cite{macina2025mathtutorbench, li2024automated}. To address this limitation, recent approaches increasingly rely on LLM-driven synthetic data generation for essay feedback \cite{li2023distilling, do2025radme}.

\begin{figure*}[t]
    \centering
    \includegraphics[width=\linewidth, height=0.3\linewidth]{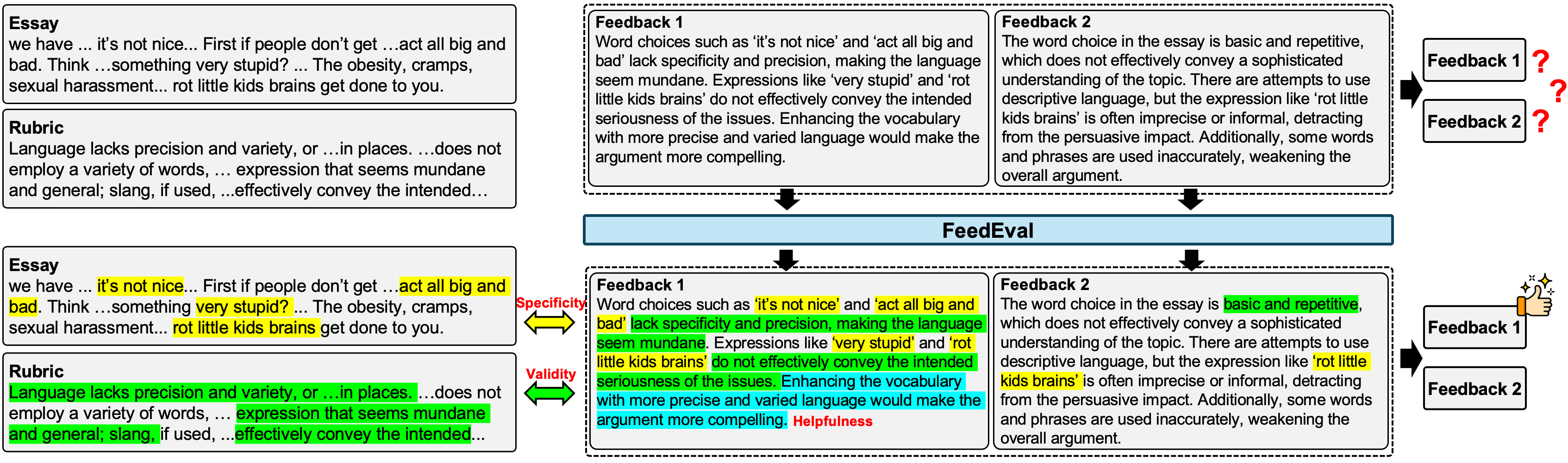}
    \caption{FeedEval evaluates the quality of multiple feedback candidates for the same essay by assessing how well they {\sethlcolor{yellow!70}\hl{reference the essay}}, {\sethlcolor{green!70}\hl{align with the rubric}}, and {\sethlcolor{myblue}\hl{provide actionable revision suggestions}}.}
    \label{fig1}
\end{figure*}

In essay scoring with feedback generation, the quality of feedback labels is crucial for training models to predict scores accurately and generate pedagogically useful feedback. However, prior approaches use LLM-generated feedback as labels without explicit quality validation. As shown in Figure \ref{fig1} (Feedback 2), using LLM-generated feedback without quality evaluation can introduce low-quality outputs that do not refer to the content of the essay, deviate from the scoring rubric, or provide limited actionable guidance, ultimately degrading downstream applications such as model training. To address this issue, we propose \textbf{FeedEval} (\textbf{Feed}back \textbf{Eval}uation), an LLM-based framework for evaluating the pedagogical quality of multiple LLM-generated essay feedback candidates and identifying high-quality feedback. 

Through FeedEval, we fine-tune LLMs to assess feedback quality along three \textbf{pedagogically grounded dimensions} — specificity \cite{hattie2007power, shute2008focus}, helpfulness \cite{steiss2024comparing}, and validity \cite{black2009developing}. This design yields evaluations that closely align with human teachers’ judgments of feedback quality, which we refer to as \textbf{pedagogically aligned evaluation} and confirm through multiple experiments involving experts in the educational domain. 

As shown in Figure \ref{fig1} (Feedback 1), FeedEval filters high-quality feedback closely tailored to both the essay content and the scoring rubric, which can further serve as reliable supervision for downstream tasks, including essay evaluation model training. Specifically, we construct or adapt dimension-relevant datasets to train the LLM evaluators for each dimension. Then, we prompt an LLM to generate multiple feedback candidates for each essay via temperature sampling and evaluate them using the dimension-specific LLM evaluators. Based on the resulting scores, FeedEval selects high-quality feedback from multiple candidates.

We conduct our experiments on the ASAP++ dataset of student essays with human-annotated multi-trait scores. We first validate the alignment of FeedEval with educational experts in judging the quality of LLM-generated essay feedback. We then evaluate the effectiveness of FeedEval by comparing the essay scoring accuracy of LLMs trained on high- and low-quality feedback filtered by FeedEval. In addition, we examine the pedagogical usefulness of the high-quality feedback through essay revision experiments and human evaluations. Our extensive experiments show that FeedEval achieves close alignment with human expert judgments, that training with FeedEval-filtered high-quality feedback leads to more accurate essay scoring, and that the resulting feedback is more pedagogically helpful than its low-quality counterpart. 


To summarize, our main contributions are as follows: (1) we propose \textbf{FeedEval}, an LLM-based framework that evaluates essay feedback along three pedagogical dimensions—specificity, helpfulness, and validity—and release \textbf{SpecEval}, the first dataset for training essay specificity evaluation models, (2) we demonstrate that FeedEval exhibits close agreement with expert judgments in essay feedback evaluation via human expert evaluations, and (3) through extensive experiments, we show that FeedEval-filtered high-quality feedback leads to more accurate essay scoring and is pedagogically more meaningful than low-quality feedback.

\section{Related Work}
\label{related_work}

\subsection{LLM-based Essay Assessment}



Recent research on automated essay assessment has increasingly focused on leveraging LLMs, including conventional language models (e.g., BERT), for essay scoring and feedback generation \cite{yang2020enhancing, wang, do_cross_prompt, do2024, lee, li2025graph}. However, jointly generating essay scores and feedback typically requires expert-written, pedagogically grounded feedback, which is costly and impractical in real-world settings \cite{li2024automated}. To address this limitation, recent approaches use LLMs to generate essay feedback or rationales \cite{chu2025rationale, do2025radme, li2023distilling}, and leverage them together with human-annotated scores as supervision for training essay assessment models.

Despite these advances, most existing methods implicitly assume that LLM-generated feedback is pedagogically reliable and do not explicitly assess its quality. As a result, low-quality feedback (e.g., overly generic or misleading feedback) may be included in the training data, potentially degrading model performance. In contrast, our work systematically evaluates the pedagogical quality of LLM-generated feedback and filters high-quality feedback, which can be used to develop more reliable LLM-based essay assessment models.

\subsection{Evaluation of LLM-generated Feedback}
Evaluating the quality of educational feedback is a longstanding challenge in the learning sciences area \cite{stahl2024exploring, meyer2024using, behzad2024assessing}. Prior work emphasizes that effective feedback should be grounded in students’ work, rubric-aligned, and provide actionable guidance \cite{hattie2007power, shute2008focus, black2009developing, steiss2024comparing}. However, these criteria are inherently open-ended and context-dependent, requiring expert pedagogical judgment, which makes high-quality feedback annotation costly, labor-intensive, and hard to scale \cite{macina2025mathtutorbench, li2024automated}. As a result, reference-based metrics such as BLEU \cite{papineni2002bleu} and BERTScore \cite{zhang2019bertscore} are ill-suited for evaluating the quality of LLM-generated feedback, while LLM-as-a-judge approaches show limited agreement with human experts and raise pedagogical validity concerns \cite{li2025generation, liang2024mathchat, maurya2025unifying}.

To address this gap, we propose FeedEval, a feedback evaluation framework tailored to essay assessment. FeedEval provides a scalable, pedagogy-aware proxy for evaluating and filtering LLM-generated feedback, and is validated through direct comparisons with expert teachers’ judgments.

\section{FeedEval Framework}
\label{feedeval}
\begin{figure}[t]
    \centering
    \includegraphics[width=\linewidth]{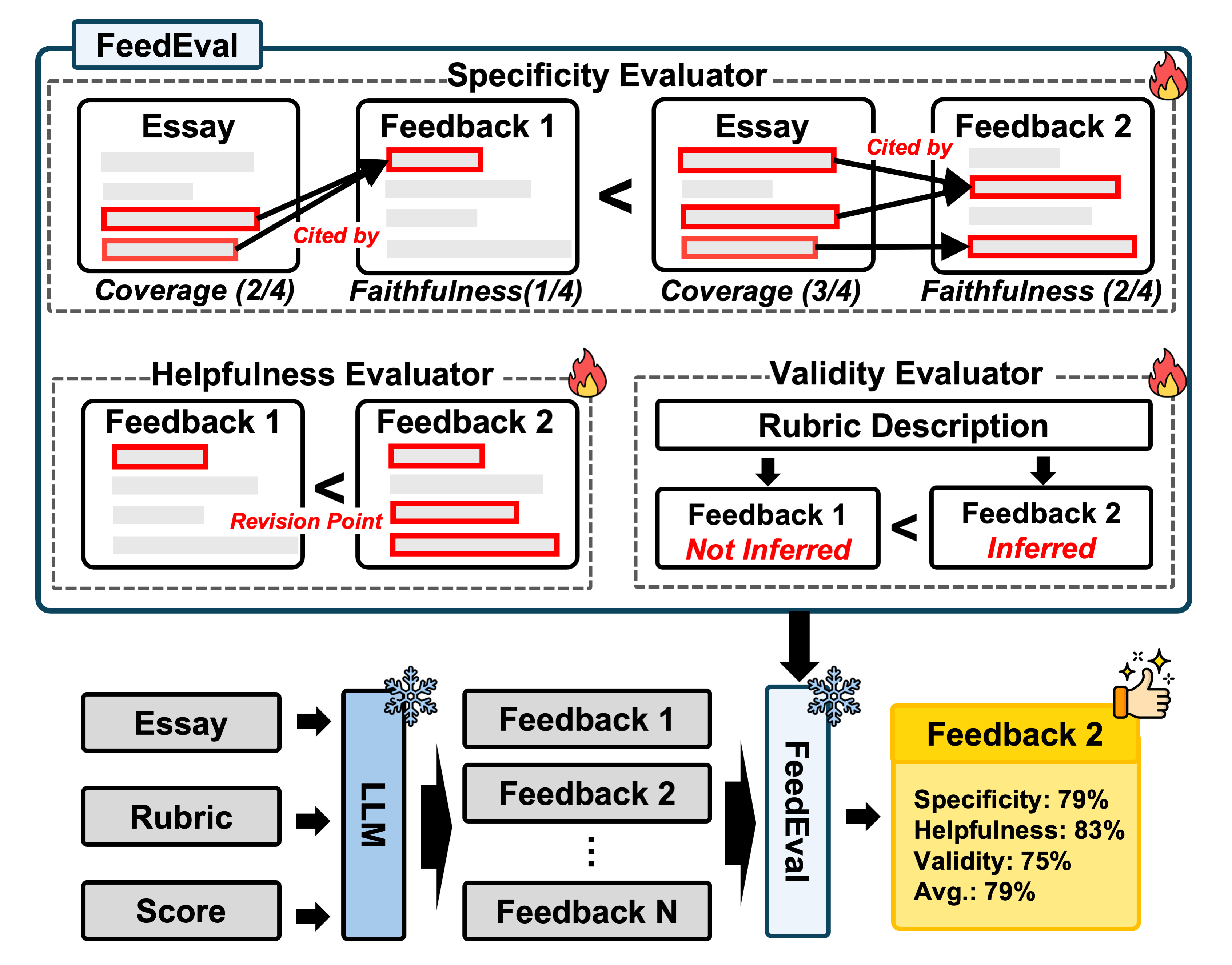}
    \caption{Overview of the proposed FeedEval framework. FeedEval consists of three evaluators, each trained on a dataset corresponding to a specific evaluation dimension. Given multiple feedback candidates generated by an LLM, FeedEval evaluates their quality along three dimensions and selects the highest-quality feedback.}
    \label{fig2}
\end{figure}

We propose a new framework, FeedEval, for evaluating LLM-generated essay feedback across multiple traits and filtering high-quality feedback from multiple candidates. Figure \ref{fig2} illustrates FeedEval in detail. Its components are explained below. 

\subsection{Evaluation Dimensions}
Extant literature on educational feedback underscores several multifaceted components essential for instructional efficacy. Specifically, effective feedback must facilitate feed-up by clarifying learning objectives, feed-back through precise diagnostic assessments of current performance, and feed-forward by offering strategic pathways for improvement \cite{hattie2007power}. Furthermore, integrating the five quality dimensions established in prior research \cite{scarlatos2024improving, steiss2024comparing, black2009developing}, feedback should (1) maintain non-revelatory guidance to preserve cognitive challenge (revealing), (2) ensure factual accuracy for reliable support (correctness), (3) provide diagnostic precision in evaluating learner levels (diagnosticity), (4) offer actionable scaffolding for revision (guidance), and (5) employ an encouraging tone to foster motivation (encouragement).

Synthesizing these theoretical foundations, this study evaluates LLM-generated essay feedback across three pedagogically grounded dimensions:

\begin{itemize}[leftmargin=0.7em, itemsep=0pt, topsep=0pt]
    \item \textbf{Specificity}: Feedback includes explicit references to relevant parts of the student's essay.
    \item \textbf{Helpfulness}: Feedback provides actionable guidance that supports the student's improvement.
    \item \textbf{Validity}: Feedback accurately diagnoses the quality of the student's essay based on the rubric score descriptions.
\end{itemize}

Regarding the feedback dimensions proposed in \citealt{hattie2007power}, we explicitly omit the feed-up component. In essay scoring, learning targets are inherently defined by the scoring rubrics and remain invariant across samples; thus, reiterating these objectives offers marginal instructional value while potentially inflating feedback length and compromising the efficiency of LLM generation. Instead, our framework concentrates on feed-back—operationalized through specificity and validity—and feed-forward, captured by helpfulness, to provide targeted diagnostic and improvement-oriented signals. Furthermore, we refine the five feedback quality dimensions established in prior work \cite{scarlatos2024improving, steiss2024comparing, black2009developing} by excluding the revealing aspect, as essay tasks lack a singular, predefined correct answer. The remaining four attributes are systematically mapped onto our evaluation criteria: correctness and diagnosticity correspond to specificity and validity, respectively, while guidance and encouragement are integrated into helpfulness, as both are essential for scaffolding the learner's iterative refinement process.

\subsection{Evaluation Pipeline}
We employ dimension-specific LLM evaluators trained to assess each feedback dimension. As shown in Algorithm \ref{feedback_selection_algorithm}, FeedEval evaluates multiple feedback candidates across traits and selects the highest-quality one per trait. For each dimension, scores are normalized across feedback candidates using a softmax function. The following sections describe how each evaluator is constructed.

\begin{algorithm}[t]
\small
\caption{FeedEval-based Feedback Selection}
\label{feedback_selection_algorithm}
\begin{algorithmic}[1]
\State \textbf{Input:} Essay $e$; Traits $\mathcal{T}$; Feedback candidates $\{\mathcal{F}_t\}_{t\in\mathcal{T}}$ with $\mathcal{F}_t=\{f_t^{(1)},\ldots,f_t^{(N)}\}$; Rubric score descriptions $\{r_t\}_{t\in\mathcal{T}}$
\State \textbf{Output:} Selected feedback $\mathcal{F}^\star$
\State $\mathcal{F}^\star \gets \emptyset$
\For{$t \in \mathcal{T}$}
    \For{$i = 1$ to $N$}
        \State $s_{\text{spec}}^{(i)} \gets \text{Spec}(e,f_t^{(i)})$
        \State $s_{\text{help}}^{(i)} \gets \text{Help}(e,f_t^{(i)})$
        \State $s_{\text{valid}}^{(i)} \gets \text{Valid}(r_t,f_t^{(i)})$
    \EndFor
    \State $\tilde{\mathbf{s}}_{\text{spec}} \gets \text{Softmax}\!\left(\{s_{\text{spec}}^{(i)}\}_{i=1}^{N}\right)$
    \State $\tilde{\mathbf{s}}_{\text{help}} \gets \text{Softmax}\!\left(\{s_{\text{help}}^{(i)}\}_{i=1}^{N}\right)$
    \State $\tilde{\mathbf{s}}_{\text{valid}} \gets \text{Softmax}\!\left(\{s_{\text{valid}}^{(i)}\}_{i=1}^{N}\right)$
    \State $i^\star \gets \arg\max_{i \in \{1,\ldots,N\}}
    \frac{1}{3}\Big(
        \tilde{\mathbf{s}}_{\text{spec}}^{(i)} +
        \tilde{\mathbf{s}}_{\text{help}}^{(i)} +
        \tilde{\mathbf{s}}_{\text{valid}}^{(i)}
    \Big)$
    \State $\mathcal{F}^\star \gets \mathcal{F}^\star \cup \{f_t^{(i^\star)}\}$
\EndFor
\State \Return $\mathcal{F}^\star$
\end{algorithmic}
\end{algorithm}


\subsubsection{Specificity Evaluator} \label{specificity_evaluator_detail}

The specificity evaluator measures how faithfully and widely feedback references a student’s essay. Given that no existing dataset explicitly targets the specificity of essay feedback, we construct a new 41K dataset (\textbf{SpecEval}\footnote{We release the dataset for future research.}) using GPT-4o, leveraging its strong capabilities in document comparison and fine-grained textual alignment \cite{chu2025think}.
For each essay, we generate three feedback variants using different prompt designs (see Appendix \ref{prompt_for_feedback_generation}), resulting in three essay–feedback pairs. Given each pair, the LLM extracts essay segments directly quoted in the feedback. We then compute two metrics: (1) the proportion of feedback sentences that reference the essay (faithfulness) and (2) the proportion of essay sentences referenced by the feedback (coverage). Specificity is defined as the F1 score of these two metrics. Using these scores, we construct a chosen–rejected pairwise dataset by ranking feedback candidates for the same essay. 

To build the specificity evaluator for selecting the superior feedback from a pair, we train an LLM-based reward model on the SpecEval dataset using a binary ranking loss following \citet{ouyang2022training}: 
\begin{equation}
\small
\mathcal{L}_{\text{rank}}
= -\frac{1}{N}\sum_{i=1}^{N}\log\sigma\!\left(
r_\theta(e_i,f_i^{+}) - r_\theta(e_i,f_i^{-}) - m
\right)    
\end{equation}

where $r_{\theta}(e, f)$ is the scalar score for an essay $e$ and feedback $f$, $f^{+}$ and $f^{-}$ denote the chosen and rejected feedback, respectively, $m$ enforces a margin between them \cite{macina2025mathtutorbench}, and $\sigma$ denotes the sigmoid function.

\subsubsection{Helpfulness Evaluator}

The helpfulness evaluator assesses how well an LLM-generated feedback provides actionable revision points—that is, concrete suggestions that a student can directly apply to improve an essay. 

To capture this notion, we construct a 14K pairwise dataset by adapting prior feedback datasets \cite{recipe4u, feat} and human-written feedback samples from the ASAP++ dataset to our task setting. Specifically, we reformat these datasets into chosen–rejected pairs based on whether the feedback offers clear, actionable guidance for revision, and use them to train the helpfulness evaluator (see Appendix \ref{helpfulness_dataset_detail} for details of the datasets and reformatting strategy). Following the same training strategy as the specificity evaluator, we train an LLM-based reward model using a binary ranking loss \cite{ouyang2022training}. 

\subsubsection{Validity Evaluator}

The validity evaluator assesses whether feedback accurately diagnoses a student’s essay with respect to the rubric score descriptions. We formulate validity evaluation as a natural language inference (NLI) task, assuming that high-validity feedback should be readily inferable from the rubric score descriptions corresponding to an essay’s score. More specifically, we treat the rubric score descriptions as the premise and the feedback as the hypothesis.
To train the validity evaluator, we use the Prometheus dataset\footnote{\url{https://huggingface.co/datasets/prometheus-eval/Feedback-Collection}}, designed to train LLMs that align closely with human judgments for evaluating open-ended responses based on rubric guidelines. The dataset is a synthetic dataset generated by GPT-4 including open-ended responses, scoring rubrics, scores, and corresponding feedback. We reformulate this dataset into a validity-focused NLI task by pairing rubric score descriptions with feedback: for each response, the rubric score descriptions corresponding to the evaluated score is treated as the premise and the feedback as the hypothesis, labeled as \textit{entailment}, while pairing the same feedback with rubric descriptions from a randomly selected different score level is labeled as \textit{contradiction}. This process yields a 99K paired NLI-style dataset, enabling the model to learn whether feedback accurately reflects the rubric-defined level of an essay.

We train an LLM using the loss: 
\begin{equation}
\small
\mathcal{L}_{\text{NLI}}
= -\frac{1}{N}\sum_{i=1}^{N}
\log p_\theta\!\left(
y_i \mid rubric^{pre}_i, feedback^{hyp}_i
\right)
\end{equation}

where $p_{\theta}(y \mid rubric^{pre}, feedback^{hyp})$ denotes the probability of generating the NLI label $y\in\{\textit{entailment}, \textit{contradiction}\}$, conditioned on the $rubric^{pre}$ and the $feedback^{hyp}$. The probability of generating \textit{entailment} is used as the initial score prior to normalization.



\section{Experimental Setup}
\label{experimental_setup}
We conduct extensive experiments to examine (1) the alignment between FeedEval instantiated with different LLMs and human experts in evaluating the quality of essay feedback, (2) the impact of FeedEval-assessed feedback quality on training LLMs for essay scoring, and (3) the effectiveness of FeedEval-assessed feedback quality in guiding essay revision by small LLMs. Our study is guided by the following three research questions:

\begin{itemize}
    \item \textbf{RQ1.} How well does FeedEval, instantiated with different LLMs, align with human expert judgments when evaluating essay feedback across specificity, helpfulness, and validity?
  
   \item \textbf{RQ2.} How does feedback quality assessed by FeedEval influence the training of LLMs for essay scoring?
  
   \item \textbf{RQ3.} How does feedback quality assessed by FeedEval affect essay revision by small LLMs?
\end{itemize}
  
  
  

\subsection{Datasets}
To evaluate FeedEval’s ability to assess essay feedback quality, we use the ASAP++ dataset \cite{mathias2018++}, an enhanced version of the ASAP dataset \cite{asap}, that provides human-annotated multi-trait scores for English essays across six prompts. As shown in Table \ref{asap_data_description}, different prompts are evaluated using different writing traits. We exclude the original ASAP dataset because it reports only aggregated scores from two annotators, making it difficult to align scores with explicit rubric descriptions.

\begin{table}[htbp]
\centering
\scriptsize
\setlength{\tabcolsep}{2pt}
\begin{tabular}{|>{\centering\arraybackslash}m{1cm}|c|c|c|}
    \toprule
    \textbf{Dataset} & \textbf{Prompt} & \textbf{\# Essays} & \textbf{Traits} \\
    \midrule
    \multirow{6}{*}{\rotatebox{90}{\textbf{ASAP++}}} 
    & 1 & 1783 & Over, Cont, WC, Org, SF, Conv \\
    & 2 & 1800 & Over, Cont, WC, Org, SF, Conv \\
    & 3 & 1726 & Over, Cont, PA, Nar, Lang \\
    & 4 & 1772 & Over, Cont, PA, Nar, Lang \\
    & 5 & 1805 & Over, Cont, PA, Nar, Lang \\
    & 6 & 1800 & Over, Cont, PA, Nar, Lang \\ 
    \bottomrule
\end{tabular}
\caption{Prompt-trait composition of the ASAP++ dataset. Traits include Overall (Over), Content (Cont), Word Choice (WC), Organization (Org), Sentence Fluency (SF), Conventions (Conv), Prompt Adherence (PA), Narrativity (Nar), and Language (Lang).}
\label{asap_data_description}
\end{table}

\subsection{Human Expert Alignment of FeedEval} \label{expert_alignment_feedeval}
We evaluate the alignment between FeedEval and educational experts in judging feedback quality across three dimensions. FeedEval is instantiated with 3B-scale LLM backbones (approximately 3–4B parameters) and fine-tuned on dimension-specific datasets to analyze expert alignment. For comparison, we also include GPT-5.1 and Gemini-2.5-Pro using LLM-as-a-judge prompting, representing widely adopted LLM-based evaluation approaches \cite{li2025generation, zheng2023judging}. We employ four prompting strategies: zero-shot, zero-shot with Chain-of-Thought (CoT), few-shot, and few-shot with CoT. Detailed prompt templates are provided in Appendix \ref{llm_as_a_judge_prompt_template}.



Alignment is assessed via pairwise comparisons constructed from GPT-5.1-generated feedback\footnote{A different model version is used for evaluation to avoid overlap with GPT-4o, which was used to construct the specificity dataset (SpecEval).}. Following prior work \cite{li2023distilling, do2025radme, chu2025rationale}, we design three prompting settings that vary the inclusion of essays, prompts, excerpts, human-annotated scores, and rubric descriptions (see Appendix \ref{prompt_for_feedback_generation}). Three educational experts selected the better feedback per dimension, and alignment is measured by whether the feedback with the higher FeedEval score matches expert preferences\footnote{Human evaluation details are provided in Appendix \ref{detail_of_pairwise_comparison}.}. We report accuracy and F1 scores for the resulting pairwise rankings.

\subsection{Feedback Quality Evaluation}

To examine the feedback quality evaluated by FeedEval, we prompt GPT-5.1 to generate eight feedback candidates per essay and trait using temperature sampling (temperature = 0.7). We then select feedback with the highest or lowest average FeedEval scores across specificity, helpfulness, and validity, referred to as high- and low-quality feedback, respectively. The Overall trait is excluded due to the absence of rubric descriptions. To further assess FeedEval’s filtering effectiveness, we compare these results with feedback filtered by GPT-5.1, which selects feedback of the highest- or lowest-quality from the candidate set without relying on FeedEval scores. We adopt few-shot prompting with CoT as it demonstrates the highest alignment with human experts among the evaluated proprietary LLM configurations (see Table \ref{human_alignment_table}).

\subsection{Impact of Feedback Quality on Essay Scoring} \label{experiment_essay_scoring}
Since high-quality LLM-generated rationales effectively supervise smaller LLMs’ reasoning \cite{kang2023knowledge}, we train 8B-scale LLMs to jointly generate multi-trait scores and feedback\footnote{Implementation details are provided in Appendix~\ref{implementation_detail}.}. Outputs follow a structured JSON format, where each trait contains a score and feedback (e.g., \texttt{\{\textbf{content:} \{score:3.0, feedback:...\}, \textbf{word choice:} \{score:2.0, feedback:...\},...\}}), while feedback for the Overall trait is set to ``\texttt{NAN}''. This design supports generating long outputs and enables score prediction followed by feedback generation. Scoring performance is evaluated using quadratic weighted kappa (QWK) \cite{cohen1968weighted} with 5-fold cross-validation, comparing models trained with high- and low-quality feedback labels.

\subsection{Impact of Feedback Quality on Essay Revision}
We investigate the impact of feedback quality on essay revision through two approaches. First, following prior work \cite{nair2024closing, llmtutor}, we use small-sized LLMs as student simulators to revise human-written essays guided by feedback of varying quality, and measure revision gains using a fine-tuned essay scoring model. This design is motivated by evidence that small LLMs struggle to produce high-quality text compared to larger models \cite{song2025learning, eldan2023tinystories}, making them a reasonable proxy for learners who benefit from feedback. Although validation with real students is ultimately necessary, this controlled setting enables scalable and reproducible evaluation of feedback effectiveness without the ethical and privacy constraints associated with studies involving human learners \cite{macina2025mathtutorbench}. Second, we conduct human evaluations in which domain experts compare feedback identified as high- or low-quality by FeedEval to assess its pedagogical usefulness.

\section{Experimental Results}
\label{main_experiment}
\subsection{Human Expert Alignment of FeedEval (RQ1)}

\subsubsection{Alignment Across Different LLMs}

\begin{table}[h!]
\centering
\renewcommand{\arraystretch}{0.9}
\resizebox{\columnwidth}{!}{%
\begin{tabular}{|l|cc|cc|cc|}
\toprule
\multicolumn{1}{|c|}{\multirow{2}{*}{\textbf{Model}}}
& \multicolumn{2}{c|}{\textbf{Specificity}} 
& \multicolumn{2}{c|}{\textbf{Helpfulness}} 
& \multicolumn{2}{c|}{\textbf{Validity}} \\ \cmidrule{2-7}
& \textbf{Acc.}  & \textbf{F1}    
& \textbf{Acc.}  & \textbf{F1}    
& \textbf{Acc.}  & \textbf{F1}    \\ \midrule
GPT (zero-shot)        & 0.729          & 0.833          & 0.584          & 0.697          & 0.640          & 0.445          \\
GPT (zero-shot+CoT)        & 0.735          & 0.831          & 0.590          & 0.699          & 0.656          & 0.457          \\
GPT (few-shot)        & 0.742          & 0.847          & 0.599          & 0.706          & 0.661          & 0.462          \\
GPT (few-shot+CoT)        & 0.751          & 0.859          & 0.615          & 0.720          & 0.682          & 0.479          \\
Gemini (zero-shot) & 0.757          & 0.845          & 0.556          & 0.622          & 0.613          & 0.417          \\  
Gemini (zero-shot+CoT) & 0.761          & 0.859          & 0.567          & 0.639          & 0.624          & 0.428          \\  
Gemini (few-shot) & 0.769          & 0.856          & 0.571          & 0.643          & 0.638          & 0.439          \\  
Gemini (few-shot+CoT) & 0.783          & 0.869          & 0.597          & 0.659          & 0.653          & 0.458          \\  \midrule
Llama3-3B-Inst. (3B-scale)   & \underline{0.820}           & \underline{0.880}           & \underline{0.864}          & \underline{0.912}          & \textbf{0.835} & \textbf{0.709} \\
Qwen2-3B-Inst. (3B-scale)   & 0.807          & 0.870           & 0.755          & 0.824          & \underline{0.833}          & \underline{0.703}          \\
Phi-3-Mini (3B-scale)    & 0.811          & 0.860           & \textbf{0.871} & \textbf{0.920} & 0.820          & 0.700          \\ 
Gemma3-Inst. (3B-scale)  & \textbf{0.832} & \textbf{0.893} & 0.853          & 0.895          & 0.822          & 0.702          \\ \bottomrule
\end{tabular}%
}
\caption{Human alignment of FeedEval across three feedback dimensions (pairwise Acc./F1). All 3B-scale models are fine-tuned. GPT and Gemini refer to GPT-5.1 and Gemini-2.5-Pro respectively. The best performances are shown in \textbf{bold}, and the second-best are \underline{underlined}.}
\label{human_alignment_table}
\end{table}

Table \ref{human_alignment_table} reports the alignment between FeedEval and human experts across the three feedback quality dimensions for different LLM backbones. \textbf{3B-scale models trained on dimension-specific datasets achieve the strong alignment, consistently attaining accuracies above 75\% and F1 scores above 70\% across the three dimensions}. These fine-tuned models outperform larger frozen models, including GPT-5.1 and Gemini-2.5-Pro, highlighting the effectiveness of dimension-specific fine-tuning for feedback quality evaluation. Among the frozen proprietary models, GPT-5.1—when configured with few-shot prompting and CoT—demonstrates the highest overall alignment with human expert annotations. 

Based on its consistent strong performance across all of the dimensions, we adopt Llama3-3B-Instruct as the final FeedEval backbone. Cohen’s Kappa of the backbone shows substantial agreement for helpfulness (0.62) and moderate agreement for specificity\footnote{The GPT-4o-based specificity computation used for SpecEval dataset construction shows high human alignment (accuracy: 89.2\%, F1: 92.0\%, Cohen’s Kappa: 0.72); however, we use the fine-tuned 3B model for efficiency.} (0.52) and validity (0.59).

To investigate potential model-dependency and biases stemming from the exclusive use of the GPT family (GPT-4o for SpecEval construction and GPT-5.1 for essay feedback generation), we additionally construct an alternative SpecEval dataset using Llama3-70B. We subsequently train evaluators to compare its alignment with expert annotations as well as with the evaluators trained on GPT-4o-based SpecEval. As a result, we observe that the evaluator trained on the Llama3-70B-synthesized data demonstrate comparable human–LLM agreement in terms of accuracy and F1 score to its GPT-4o-based counterpart. Furthermore, the evaluators trained on the GPT-4o-generated and Llama3-70B-generated SpecEval datasets achieve high agreement scores (exceeding 0.8 in both metrics). These results suggest that both models capture consistent pedagogical characteristics of essay feedback, implying that the choice of generator LLM does not significantly bias the resulting evaluator's performance. Detailed results are shown in Appendix \ref{llama_speceval}.



\subsubsection{Impact of Feedback-related Knowledge}

\begin{table}[H]
\centering
\resizebox{\columnwidth}{!}{%
\begin{tabular}{|cc|ccc|}
\toprule
\multicolumn{2}{|c|}{\textbf{Knowledge Configuration}} 
& \multicolumn{3}{c|}{\textbf{Accuracy/F1-score}} \\ \midrule
\rowcolor{white}
\textbf{Task-related} & \textbf{Feedback-related} 
& \textbf{Specificity} & \textbf{Helpfulness} & \textbf{Validity} \\ \midrule
\rowcolor{white}
\xmark & \xmark 
& 0.404 / 0.489 & 0.262 / 0.000 & 0.513 / 0.282 \\
\rowcolor[HTML]{EFEFEF}
\cmark & \xmark 
& 0.644 / 0.748 & 0.656 / 0.768 & 0.671 / 0.417 \\
\rowcolor[HTML]{C0C0C0}
\cmark & \cmark 
& 0.820 / 0.880 & 0.864 / 0.912 & 0.835 / 0.709 \\
\bottomrule
\end{tabular}%
}
\caption{Alignment of Llama-based fine-tuned models with expert judgments across knowledge configuration.}
\label{table_1}
\end{table}

Given our goal of enabling LLMs to evaluate feedback quality, we examine the role of feedback-specific knowledge in aligning LLM judgments with those of human experts. To this end, we compare three settings: no fine-tuning, fine-tuning on generic task-oriented data, and fine-tuning on feedback-specific data constructed in this study. For the generic task-oriented fine-tuning, we train specificity and helpfulness evaluators using a human-preference reward dataset\footnote{Anthropic RLHF \cite{bai2022training}, \url{https://huggingface.co/datasets/Anthropic/hh-rlhf}.}, and train the validity evaluator on the MNLI dataset\footnote{MNLI \cite{mnli}, \url{https://huggingface.co/datasets/nyu-mll/multi_nli}.}, formulated as an NLI task. As shown in Table \ref{table_1}, \textbf{while the generic task-oriented fine-tuning improves alignment, fine-tuning with our curated datasets yields further gains, demonstrating the importance of feedback-specific knowledge for reliable LLM-based feedback assessment}.

\subsubsection{Analysis of LLM-generated Feedback}

\begin{table}[h]
\centering
\scriptsize

\begin{tabular}{|cc|ccc|}
\toprule
\textbf{Score} & \textbf{Rubric} & \textbf{Specificity} & \textbf{Helpfulness} & \textbf{Validity} \\ \midrule
\cmark & \cmark & \textbf{56.68} & \underline{40.71}  & \textbf{63.72} \\
\cmark & \xmark & 2.22 & 9.48  & \underline{27.31} \\
\xmark & \cmark & \underline{41.10} & \textbf{49.81} & 8.97  \\ \bottomrule
\end{tabular}%
\caption{Proportion (\%) of feedback-generation methods that achieve the highest score per dimension for each essay. The best results are shown in \textbf{bold}, and the second-best are \underline{underlined}.}
\label{feedback_analysis}
\end{table}

We analyze feedback quality across three prompting strategies specified in Section \ref{expert_alignment_feedeval} by measuring how often each method achieves the highest FeedEval score per essay across specificity, helpfulness, and validity. As shown in Table \ref{feedback_analysis}, prompting with both human-annotated scores and rubric score descriptions most consistently yields high-quality feedback, particularly for specificity and validity, underscoring the importance of score–rubric integration. Rubric-only prompting achieves the best performance for helpfulness and the second-best for specificity, highlighting the crucial role of rubric information. Accordingly, we adopt the score–rubric prompting strategy to generate feedback candidates in subsequent experiments.

\subsection{Impact of Feedback Quality on Essay Scoring (RQ2)} \label{feedback_quality_impact_on_scoring}

\subsubsection{Essay Scoring Performance} \label{scoring_performance}

\begin{table*}[]
\centering
\resizebox{\textwidth}{!}{%
\begin{tabular}{|ccc|ccccccccc|c|}
\toprule
 & & &
  \multicolumn{9}{c|}{\textbf{Traits (Prediction Order: $\leftarrow$)}} & \\ \midrule
\multicolumn{1}{|c|}{\textbf{LLM}} &
  \multicolumn{1}{c|}{\textbf{Assessment}} &
  \textbf{Feedback Quality} &
  \textbf{Over} &
  \textbf{Cont} &
  \textbf{PA} &
  \textbf{Lang} &
  \textbf{Nar} &
  \textbf{Org} &
  \textbf{Conv} &
  \textbf{WC} &
  \textbf{SF} &
  \textbf{Avg$\uparrow$ (SD$\downarrow$)} \\ \midrule

\multicolumn{1}{|c|}{} &
  \multicolumn{1}{c|}{Score Only} &
  \xmark &
  \textbf{0.476} & 0.546 & 0.585 & 0.540 & 0.580 & \textbf{0.580} & 0.482 & 0.491 & 0.476 &
  0.528 ($\pm$0.040) \\ \cmidrule{2-13}

\multicolumn{1}{|c|}{} &
  \multicolumn{1}{c|}{\multirow{2}{*}{\centering Score + Feedback}} &
  Low Quality (GPT-5.1) &
  0.445 & 0.560 & 0.603 & 0.579 & \underline{0.611} & 0.550 & \underline{0.561} &
  0.568 & 0.542 & 0.558 ($\pm$0.032) \\
\multicolumn{1}{|c|}{} &
  \multicolumn{1}{c|}{} &
  High Quality (GPT-5.1) &
  0.438 & \underline{0.592} & 0.601 & \underline{0.585} & 0.606 & 0.545 & 0.548 &
  \underline{0.575} & 0.555 & 0.561 ($\pm$0.038) \\ \cmidrule{2-13}

\multicolumn{1}{|c|}{Llama3-8B-Inst.} &
  \multicolumn{2}{c|}{\textbf{Improvement (High Quality - Low Quality)}} &
  \cellcolor{red2}-1.57\% &
  \cellcolor{blue4}+5.71\% &
  \cellcolor{red2}-0.33\% &
  \cellcolor{blue2}+1.04\% &
  \cellcolor{red2}-0.82\% &
  \cellcolor{red2}-0.91\% &
  \cellcolor{red2}-2.32\% &
  \cellcolor{blue2}+1.23\% &
  \cellcolor{blue2}+2.40\% &
  \cellcolor{blue2}+0.52\% \\ \cmidrule{2-13}

\multicolumn{1}{|c|}{} &
  \multicolumn{1}{c|}{\multirow{2}{*}{\centering Score + Feedback}} &
  Low Quality (FeedEval) &
  0.449 & 0.576 & \underline{0.605} & 0.575 & 0.602 & 0.570 & 0.556 &
  0.563 & \underline{0.558} & \underline{0.562} ($\pm$0.035) \\
\multicolumn{1}{|c|}{} &
  \multicolumn{1}{c|}{} &
  High Quality (FeedEval) &
  \underline{0.451} & \textbf{0.601} & \textbf{0.612} & \textbf{0.587} & \textbf{0.617} &
  \underline{0.575} & \textbf{0.598} & \textbf{0.598} & \textbf{0.579} &
  \textbf{0.580} ($\pm$0.037) \\ \cmidrule{2-13}

\multicolumn{1}{|c|}{} &
  \multicolumn{2}{c|}{\textbf{Improvement (High Quality - Low Quality)}} &
  \cellcolor{blue2}+0.45\% &
  \cellcolor{blue2}+4.34\% &
  \cellcolor{blue2}+1.16\% &
  \cellcolor{blue2}+2.09\% &
  \cellcolor{blue2}+2.49\% &
  \cellcolor{blue2}+0.88\% &
  \cellcolor{blue4}+7.55\% &
  \cellcolor{blue4}+6.22\% &
  \cellcolor{blue2}+3.76\% &
  \cellcolor{blue2}+3.22\% \\ \midrule

\multicolumn{1}{|c|}{} &
  \multicolumn{1}{c|}{Score Only} &
  \xmark &
  \textbf{0.712} & \underline{0.693} & 0.696 & \underline{0.677} & \underline{0.710} & \textbf{0.676} &
  0.674 & 0.681 & 0.684 & \underline{0.689} ($\pm$0.022) \\ \cmidrule{2-13}

\multicolumn{1}{|c|}{} &
  \multicolumn{1}{c|}{\multirow{2}{*}{\centering Score + Feedback}} &
  Low Quality (GPT-5.1) &
  0.649 & 0.692 & \underline{0.701} & 0.667 & 0.701 & 0.669 &
  0.682 & 0.669 & 0.675 & 0.678 ($\pm$0.026) \\
\multicolumn{1}{|c|}{} &
  \multicolumn{1}{c|}{} &
  High Quality (GPT-5.1) &
  0.657 & 0.686 & 0.697 & 0.675 & 0.699 & 0.664 &
  \underline{0.685} & \underline{0.682} & \underline{0.686} & 0.681 ($\pm$0.035) \\ \cmidrule{2-13}

\multicolumn{1}{|c|}{Qwen3-8B} &
  \multicolumn{2}{c|}{\textbf{Improvement (High Quality - Low Quality)}} &
  \cellcolor{blue2}+1.23\% &
  \cellcolor{red2}-0.87\% &
  \cellcolor{red2}-0.57\% &
  \cellcolor{blue2}+1.20\% &
  \cellcolor{red2}-0.29\% &
  \cellcolor{red2}-0.75\% &
  \cellcolor{blue2}+0.44\% &
  \cellcolor{blue2}+1.94\% &
  \cellcolor{blue2}+1.63\% &
  \cellcolor{blue2}+0.43\% \\ \cmidrule{2-13}

\multicolumn{1}{|c|}{} &
  \multicolumn{1}{c|}{\multirow{2}{*}{\centering Score + Feedback}} &
  Low Quality (FeedEval) &
  0.657 & 0.682 & 0.697 & 0.673 & 0.694 & 0.671 &
  0.656 & 0.674 & 0.684 & 0.676 ($\pm$0.023) \\
\multicolumn{1}{|c|}{} &
  \multicolumn{1}{c|}{} &
  High Quality (FeedEval) &
  \underline{0.661} & \textbf{0.699} & \textbf{0.709} & \textbf{0.683} & \textbf{0.719} &
  \underline{0.673} & \textbf{0.694} & \textbf{0.688} & \textbf{0.698} &
  \textbf{0.692} ($\pm$0.020) \\ \cmidrule{2-13}

\multicolumn{1}{|c|}{} &
  \multicolumn{2}{c|}{\textbf{Improvement (High Quality - Low Quality)}} &
  \cellcolor{blue2}+0.61\% &
  \cellcolor{blue2}+2.49\% &
  \cellcolor{blue2}+1.72\% &
  \cellcolor{blue2}+1.49\% &
  \cellcolor{blue2}+3.60\% &
  \cellcolor{blue2}+0.30\% &
  \cellcolor{blue4}+5.79\% &
  \cellcolor{blue2}+2.08\% &
  \cellcolor{blue2}+2.05\% &
  \cellcolor{blue2}+1.58\% \\ \bottomrule
\end{tabular}%
}
\caption{Average essay scoring performance across all prompts for each trait on the ASAP++ dataset. Traits are predicted from right to left ($\leftarrow$). We report five-fold averaged results with standard deviations (SD). The best performances are shown in \textbf{bold}, and the second-best are \underline{underlined} for each LLM backbone.}
\label{table_2}
\end{table*}

Table \ref{table_2} reports the essay scoring performance of Llama3-8B-Instruct and Qwen3-8B under different label configurations\footnote{Additional experiments on another dataset to examine the generalizability of FeedEval's impact on essay feedback quality evaluation are provided in the Appendix \ref{feedeval_on_sas}.}. The results show that models trained on high-quality feedback filtered by FeedEval consistently outperform those trained on low-quality feedback across all traits. In contrast, models trained on high-quality feedback filtered by GPT-5.1 do not yield consistent performance gains over their low-quality counterparts, \textbf{highlighting FeedEval's superior ability to assess and filter pedagogically helpful feedback and the effectiveness of high-quality feedback filtered by FeedEval as supervision for essay scoring}. Moreover, models trained on FeedEval-selected high-quality feedback consistently outperform those trained on GPT-5.1-selected high-quality feedback, further underscoring \textbf{FeedEval's stronger capability in identifying high-quality feedback}. 

Compared to the score-only configuration, incorporating high-quality feedback slightly degrades performance on the Overall trait. This is because the dataset does not provide rubric score descriptions for this trait, leading us to replace its feedback with ``\texttt{NAN},'' which could hinder accurate scoring. For the Organization trait, the performance gap between high- and low-quality feedback is marginal, likely because its rubric descriptions are broadly defined compared to other traits, which limits the feedback’s ability to capture structural features and leads to minimal semantic differences across feedback quality levels, as illustrated by the case study provided in Appendix \ref{appendix_case_study_feedback}. Consequently, such \textbf{unclear feedback might have slightly degraded the scoring performance of Qwen3-8B on the Organization trait relative to the score-only setting}.



Overall, the results underscore the importance of well-defined, trait-specific feedback for accurate essay scoring and demonstrate the effectiveness of FeedEval for distinguishing essay feedback quality. Given its strong performance, we adopt Qwen3-8B as the backbone for subsequent experiments.\footnote{Generating feedback before predicting scores consistently degraded performance compared to predicting scores first, as shown in the Appendix \ref{reverse_performance}.}

\subsubsection{Impact of FeedEval Dimension} \label{impact_of_dimension}

\begin{figure}[htbp]
    \centering
    \includegraphics[width=\linewidth, height=0.4\linewidth]{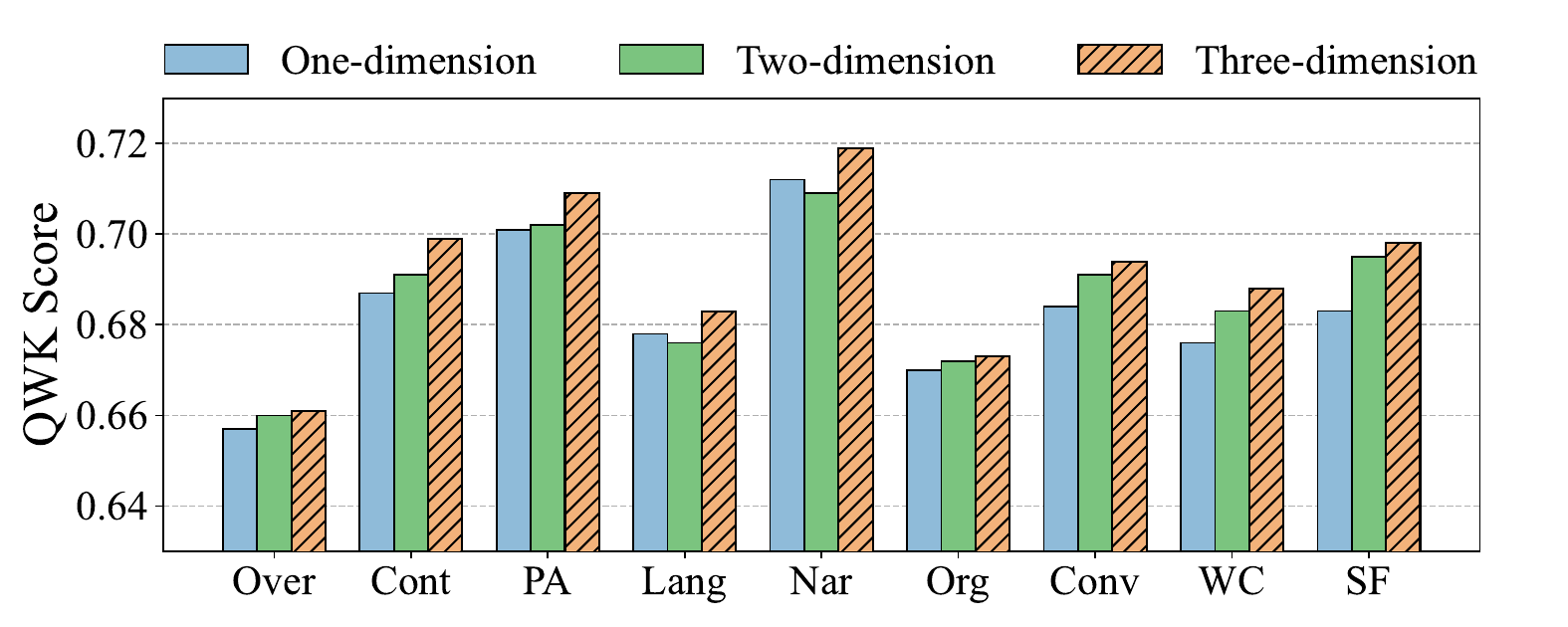}
    \caption{Average essay scoring performance across traits on ASAP++ for Qwen3-8B trained with high-quality feedback labels filtered by FeedEval using one, two, or all three dimensions.}
    \label{fig4}
\end{figure}

Figure \ref{fig4} summarizes the essay scoring performance of Qwen3-8B trained with high-quality feedback selected using different combinations of FeedEval dimensions. For one- and two-dimension settings, we average results across all possible single or pairwise combinations (see Appendix \ref{essay_scoring_dimension_config} for individual performance results). Overall, performance improves as more FeedEval dimensions are incorporated. \textbf{Using all three dimensions yields the best results across all traits}, while gains for the Overall and Organization traits remain marginal. This is likely due to the absence of feedback supervision for the Overall trait, as well as the difficulty of capturing structural features for the Organization trait, which is exacerbated by broad rubric descriptions with limited distinctions across score levels.

\subsection{Impact of Feedback Quality on Essay Revision (R3)}

\subsubsection{Essay Improvement after Revision} \label{llm_essay_revision}

\begin{figure}[htbp]
    \centering
    \includegraphics[width=\linewidth, height=0.7\linewidth]{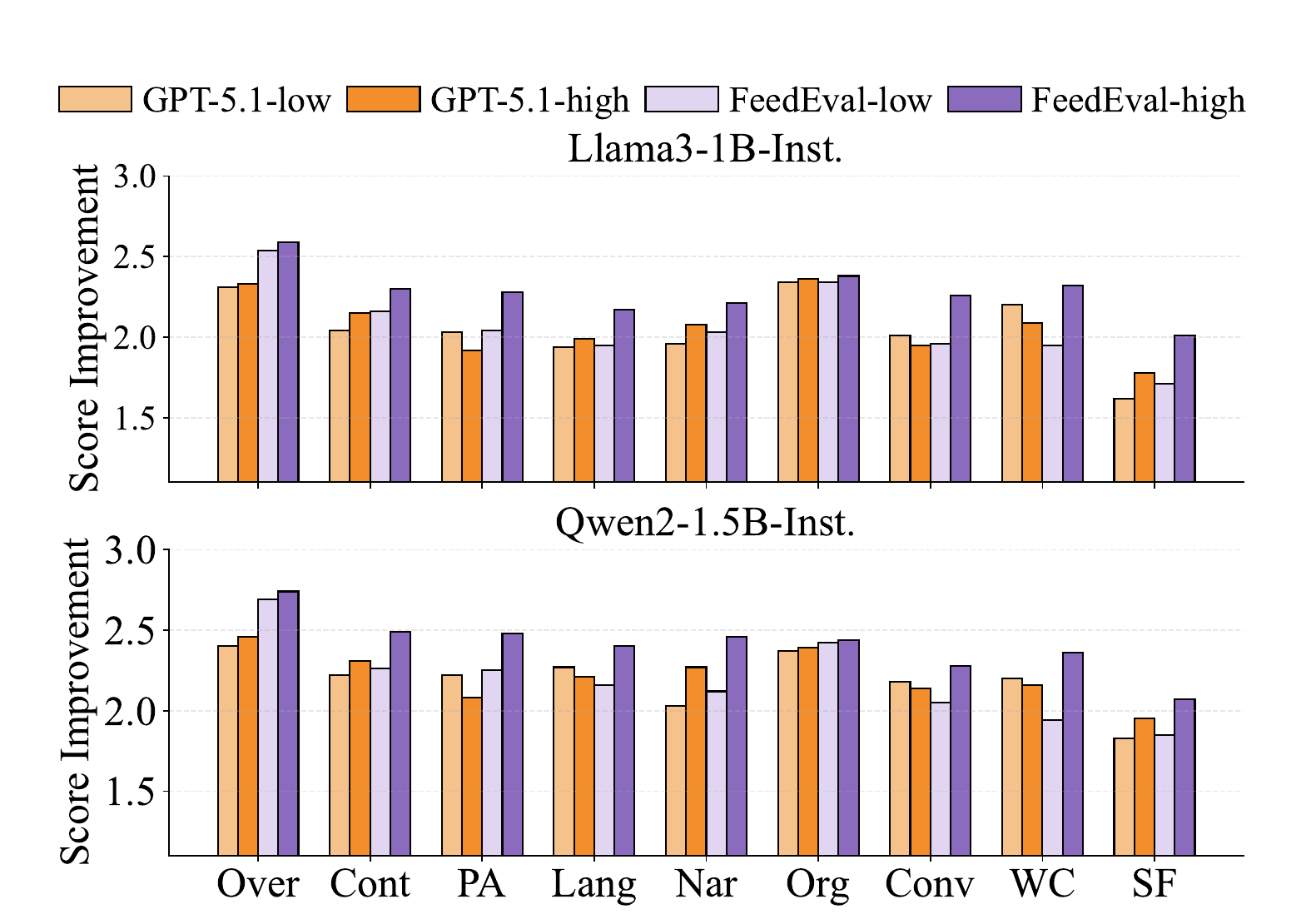}
    \caption{Average essay score improvement across traits on ASAP++ after revisions guided by feedback of high- and low-quality identified by FeedEval and GPT-5.1.}
    \label{fig5}
\end{figure}

To evaluate essay revisions by small-sized LLMs, we use the Qwen3-8B essay scoring model trained with score-only labels (Avg. QWK = 0.689; Table \ref{table_2}). Figure \ref{fig5} shows the average scores improved when Llama3-1B-Instruct and Qwen2-1.5B-Instruct revise essays using high- or low-quality feedback identified by FeedEval and GPT-5.1. 

From these results, we first observe that high-quality feedback filtered by FeedEval consistently yields larger revision gains, while improvements for the Organization trait remain marginal due to the lack of detailed rubric descriptions during feedback generation, as discussed in section \ref{feedback_quality_impact_on_scoring}. Furthermore, the revision gains from the high-quality feedback filtered by GPT-5.1 are not consistently larger than those from low-quality feedback, indicating limited capability of GPT-5.1 in distinguishing feedback quality. In addition, feedback identified as high-quality by FeedEval leads to greater revision gains than GPT-5.1-selected feedback. Overall, \textbf{these results demonstrate that FeedEval reliably identifies pedagogically high-quality feedback that enables more effective essay revisions}. Specific cases of essay feedback and revised essays are analyzed in Appendix \ref{appendix_case_study_esssay}.


\subsubsection{Human Evaluation of Feedback and Essay Revision} \label{human_evaluation_feedback_revision}

\begin{table}[H]
\centering
\resizebox{\columnwidth}{!}{%
\begin{tabular}{|
    c|
    >{\centering\arraybackslash}p{1.1cm}|
    >{\centering\arraybackslash}p{1.1cm}|
    >{\centering\arraybackslash}p{1.1cm}|
    c|
}
\toprule
\multirow{2}{*}{\textbf{Feedback}}
 & \multicolumn{3}{c|}{\textbf{Quality of Feedback (1-5)}} 
 & \multirow{2}{*}{\textbf{\begin{tabular}[c]{@{}c@{}}Quality of\\Revised Essay (\%)\end{tabular}}} \\ \cline{2-4}
 & \textbf{D1} & \textbf{D2} & \textbf{D3} & \\ \midrule
High-quality & 3.62 & 4.67 & 3.02 & 70.3 \\
Low-quality  & 1.53 & 1.30 & 2.05 & 29.7 \\ \bottomrule
\end{tabular}%
}
\caption{Human evaluation comparing feedback quality and revised essay quality.}
\label{table_4}
\end{table}

Table \ref{table_4} reports human evaluations of both feedback quality and essay revision. Three educational experts evaluated high- or low-quality feedback filtered by FeedEval for 300 essays (100 per expert), rating feedback on a 5-point Likert scale across three pedagogically grounded feedback quality dimensions proposed by \citet{steiss2024comparing}: faithfulness to essay (D1), usefulness for revision (D2), and rubric alignment (D3)\footnote{Details of each dimension are addressed in Appendix \ref{addtional_materials}}. The evaluation results show that the high-quality feedback consistently received higher scores compared to the low-quality feedback. The same experts also conducted pairwise comparisons of 300 essays revised by a small-sized LLM (Qwen2-1.5B-Instruct) using the two types of feedback, preferring essays revised with high-quality feedback significantly more often. \textbf{These results corroborate the automatic evaluation in section \ref{llm_essay_revision} and confirm that FeedEval effectively filters high-quality feedback, leading to improved downstream essay revision.}

\section{Conclusion}
\label{conclusion}
In this paper, we introduce FeedEval, a novel LLM-based framework for evaluating the pedagogical quality of LLM-generated essay feedback along three dimensions: specificity, helpfulness, and validity. We validate FeedEval’s alignment with human expert judgments and demonstrate how its evaluation scores can be used to filter high-quality feedback. Experiments on the ASAP++ dataset show that FeedEval closely matches expert evaluations and that models trained on FeedEval-filtered high-quality feedback achieve more accurate essay scoring than those trained on low-quality feedback. Moreover, essay revision experiments using small LLMs, together with human evaluations, confirm that the selected high-quality feedback is pedagogically more meaningful and effective. Finally, FeedEval consistently outperforms GPT-5.1 in distinguishing feedback quality, underscoring its effectiveness as a pedagogically aligned feedback evaluation framework and its potential to advance LLM-based automated essay assessment.

\section*{Limitations}

In this study, we identify two primary limitations. First, the essay scoring performance of the LLMs used in our experiments may be influenced by the generation order of scores and feedback, reflecting the autoregressive nature of these models. Specifically, we observe that constructing labels to generate scores before feedback leads to better scoring performance than generating feedback first. This limitation, however, is not unique to our approach and is shared by many recent LLM-based models that jointly generate predictions and rationales. Second, our study focuses exclusively on English essay writing. To evaluate the generalizability of FeedEval to broader language education settings, future work should examine its applicability to essays written in other languages.

\section*{Ethical Statement}
Our work used publicly available essay scoring benchmark datasets, including ASAP++, and did not pose any ethical concerns during experimentation. For human evaluation, we followed a well-established evaluation protocol in the literature, preventing possible ethical issues in the annotation process. Annotators were compensated at a rate approximately 30\% higher than the average U.S. minimum wage.

\section*{Scientific Artifacts}
The dataset (SpecEval) used to train the specificity evaluator was generated using GPT-4o via OpenAI’s paid API services. See Appendix \ref{specificity_evaluator_detail} for details. 

\section*{Acknowledgments}
This work was supported by the National Research Foundation of Korea(NRF) grant funded by the Korean government(MSIT) (No.RS-2022-NR068758).


\bibliography{ref}

\appendix
\label{appendix}
\section{Prompt Templates}

\subsection{Prompt Design for Feedback Generation} \label{prompt_for_feedback_generation}
We employ GPT-5.1 to generate essay feedback using multiple sources of information. For each essay, we provide the essay text, the associated prompt, and an excerpt when available. To introduce variation in the generated feedback, we additionally define three feedback-generation settings based on how human-assigned scores and rubric descriptions are incorporated.

\subsubsection{\textcolor{coral}{Score}+\textcolor{navy}{Rubric}}
In this setting, human-assigned scores and their corresponding rubric descriptions are explicitly included in the prompt. The prompt template is shown in Figure \ref{score_rubric_prompt}.

\begin{figure*}[]
\centering
\begin{tcolorbox}[
  colback=gray!5,
  colframe=black,
  fonttitle=\bfseries,
]
\small

You are a member of the English essay writing test evaluation committee. Please, evaluate the given essay using following information.

\medskip
\textbf{[Prompt]}\\
\{prompt text\}\\
\textbf{(end of [Prompt])}

\medskip
\textbf{[Excerpt]}\\
\{excerpt text\}\\
\textbf{(end of [Excerpt])}

\medskip
\textbf{[Essay]}\\
\{essay text\}\\
\textbf{(end of [Essay])}

\medskip
\textcolor{coral}{\textbf{[Scores]}}\\
\textcolor{coral}{{Narrativity: 3}}\\
\textcolor{coral}{{Language: 2}}\\
\textcolor{coral}{(\ldots)}\\
\textcolor{coral}{\textbf{(end of [Scores])}}

\medskip
\textcolor{navy}{\textbf{[Rubric descriptions]}}\\
\textcolor{navy}{\textbf{[Trait]}}\\
\textcolor{navy}{Narrativity}\\
\textcolor{navy}{\textbf{(end of [Trait])}}\\
\textcolor{navy}{The following is a rubric description in terms of the ``Narrativity'' trait.}\\
\textcolor{navy}{Score 3: The response is interesting. Appropriate use of transition and \ldots}\\
\textcolor{navy}{\textbf{[Trait]}}\\
\textcolor{navy}{Language}\\
\textcolor{navy}{(\ldots)}\\
\textcolor{navy}{\textbf{(end of [Rubric descriptions])}}

\medskip
Refer to the provided \textbf{[Prompt]}, \textbf{[Excerpt]}, \textbf{[Scores]}, and \textbf{[Rubric descriptions]} to evaluate the given essay.\\
Your task is to analyze the reason why the essay got certain scores for each trait based on the analysis of the essay.

\medskip
\textbf{[Note]}\\
I have made an effort to remove personally identifying information from the essays using the Named Entity Recognizer (NER). The relevant entities are identified in the text and then replaced with a string such as `@PERSON', `@ORGANIZATION', `@LOCATION', `@DATE', `@TIME', `@MONEY', `@PERCENT', `@CAPS' (any capitalized word) and `@NUM' (any digits). Please do not penalize the essay because of the anonymizations.\\
\textbf{(end of [Note])}

\medskip
Q. Identify specific excerpts from the [Essay] that illustrate the strengths or weaknesses highlighted in the [Rubric descriptions] for each trait. Quote or summarize the relevant parts of the essay.
Based on this analysis, rationalize the [Rubric descriptions] for each trait. If the [Rubric descriptions] for a given trait indicates that the writing is strong, provide only positive feedback. If it identifies weaknesses, provide a detailed analysis of the issue and suggest specific ways to improve it. Keep your response for each trait within three sentences, and do not include any specific scores in your analysis. Provide your answer in the following format:

\medskip
\textbf{\{``trait 1'': ``evaluation for trait 1'', ``trait 2'': ``evaluation for trait 2'', \ldots\}}

\end{tcolorbox}
\caption{Prompt template for feedback generation using both human-annotated \textcolor{coral}{\textbf{scores}} and score descriptions of the \textcolor{navy}{\textbf{rubric}}.}
\label{score_rubric_prompt}
\end{figure*}

\subsubsection{\textcolor{coral}{Score} Only}
This setting includes only the human-assigned score in the prompt, without providing the corresponding rubric description. The prompt template is shown in Figure \ref{score_only_prompt}.

\begin{figure*}[]
\centering
\begin{tcolorbox}[
  colback=gray!5,
  colframe=black,
  fonttitle=\bfseries,
]
\small

You are a member of the English essay writing test evaluation committee. Please, evaluate the given essay using following information.

\medskip
\textbf{[Prompt]}\\
\{prompt text\}\\
\textbf{(end of [Prompt])}

\medskip
\textbf{[Excerpt]}\\
\{excerpt text\}\\
\textbf{(end of [Excerpt])}

\medskip
\textbf{[Essay]}\\
\{essay text\}\\
\textbf{(end of [Essay])}

\medskip
Refer to the provided \textbf{[Prompt]} and \textbf{[Excerpt]} to evaluate the given essay. The following shows the scores of each trait provided by a human scorer.

\medskip
\textcolor{coral}{\textbf{[Scores]}}\\
\textcolor{coral}{{Narrativity: 3}}\\
\textcolor{coral}{{Language: 2}}\\
\textcolor{coral}{(\ldots)}\\
\textcolor{coral}{\textbf{(end of [Scores])}}

\medskip
Your task is to analyze the reason why the essay got certain scores for each trait.

\medskip
\textbf{[Note]}\\
I have made an effort to remove personally identifying information from the essays using the Named Entity Recognizer (NER). The relevant entities are identified in the text and then replaced with a string such as `@PERSON', `@ORGANIZATION', `@LOCATION', `@DATE', `@TIME', `@MONEY', `@PERCENT', `@CAPS' (any capitalized word) and `@NUM' (any digits). Please do not penalize the essay because of the anonymizations.\\
\textbf{(end of [Note])}

\medskip
Q. Identify specific excerpts from the [Essay] that illustrate the strengths or weaknesses for each trait. Quote or summarize the relevant parts of the essay. Based on your analysis, rationalize the score for each trait. If the writing is strong enough, provide only positive feedback. If there are some weaknesses, provide a detailed analysis of the issue and suggest specific ways to improve it. Keep your response for each trait within three sentences, and do not include any specific scores in your analysis. Provide your answer in the following format:

\medskip
\textbf{\{``trait 1'': ``evaluation for trait 1'', ``trait 2'': ``evaluation for trait 2'', \ldots\}}

\end{tcolorbox}
\caption{Prompt template for feedback generation using human-annotated \textcolor{coral}{\textbf{scores}}.}
\label{score_only_prompt}
\end{figure*}

\subsubsection{\textcolor{navy}{Rubric} Only}
This setting includes rubric descriptions covering all score ranges for each trait, without providing human-assigned scores. The LLM evaluates the essay without knowing human-annotated scores. The prompt template is shown in Figure \ref{rubric_only_prompt}.

\begin{figure*}[]
\centering
\begin{tcolorbox}[
  colback=gray!5,
  colframe=black,
  fonttitle=\bfseries,
]
\small

You are a member of the English essay writing test evaluation committee. Please, evaluate the given essay using following information.

\medskip
\textbf{[Prompt]}\\
\{prompt text\}\\
\textbf{(end of [Prompt])}

\medskip
\textbf{[Excerpt]}\\
\{excerpt text\}\\
\textbf{(end of [Excerpt])}

\medskip
\textcolor{navy}{\textbf{[Rubric guidelines]}}\\
\textcolor{navy}{\textbf{[Trait]}}\\
\textcolor{navy}{Narrativity}\\
\textcolor{navy}{\textbf{(end of [Trait])}}\\
\textcolor{navy}{\textbf{[Trait Rubric]}}\\
\textcolor{navy}{Score 0: The response is irrelevant/incorrect/incomplete.}\\
\textcolor{navy}{Score 1: The response is very uninteresting and disjointed and ...}\\
\textcolor{navy}{(\ldots)}\\
\textcolor{navy}{\textbf{(end of [Trait Rubric])}}

\medskip
\textcolor{navy}{\textbf{[Trait]}}\\
\textcolor{navy}{Language}\\
\textcolor{navy}{(\ldots)}\\
\textcolor{navy}{\textbf{(end of [Rubric guidelines])}}\\

\medskip
Refer to the provided \textbf{[Prompt]} and \textbf{[Excerpt]} to evaluate the given essay.

\medskip
\textbf{[Essay]}\\
\{essay text\}\\
\textbf{(end of [Essay])}

\medskip
\textbf{[Note]}\\
I have made an effort to remove personally identifying information from the essays using the Named Entity Recognizer (NER). The relevant entities are identified in the text and then replaced with a string such as `@PERSON', `@ORGANIZATION', `@LOCATION', `@DATE', `@TIME', `@MONEY', `@PERCENT', `@CAPS' (any capitalized word) and `@NUM' (any digits). Please do not penalize the essay because of the anonymizations.\\
\textbf{(end of [Note])}

\medskip
Q. Identify specific excerpts from the [Essay] that illustrate the strengths or weaknesses highlighted in the [Rubric guidelines] for each trait. Quote or summarize the relevant parts of the essay.
Based on your analysis, rationalize your analysis for each trait. If the writing is strong enough, provide only positive feedback. If there are some weaknesses, provide a detailed analysis of the issue and suggest specific ways to improve it. Keep your response for each trait within three sentences, and do not include any specific scores in your analysis. Provide your answer in the following format:

\medskip
\textbf{\{``trait 1'': ``evaluation for trait 1'', ``trait 2'': ``evaluation for trait 2'', \ldots\}}

\end{tcolorbox}
\caption{Prompt template for feedback generation using the \textcolor{navy}{\textbf{rubric}} guidelines.}
\label{rubric_only_prompt}
\end{figure*}

\subsection{Prompt Design for Feedback Filtering by GPT-5.1} \label{llm_as_a_judge_prompt_template}
Figure \ref{gpt_filtering_prompt} illustrates the prompt template used for filtering essay feedback quality via an LLM-as-a-judge framework, employing GPT-5.1 and Gemini-2.5-Pro. We leverage these LLMs through four distinct prompting strategies: zero-shot, zero-shot with CoT, few-shot, and few-shot with CoT. The colored segments are conditionally included based on the strategy: the black text serves as the base zero-shot prompt, while \textcolor{BrickRed}{three author-provided selection examples} are appended for the few-shot setting. The instruction ``\textcolor{Brown}{Think step by step}'' is included for CoT-based filtering. When filtering feedback based on a specific pedagogical dimension, the corresponding definition is inserted between the \textbf{[Condition]} and \textbf{(end of [Condition])} markers. On the other hand, when filtering feedback based on all three dimensions together, all the definitions are integrated into the template.

\begin{figure*}[]
\centering
\begin{tcolorbox}[
  colback=gray!5,
  colframe=black,
  fonttitle=\bfseries,
]
\small

You are an expert in filtering feedback data from multiple candidates based on specified conditions.
Read the following conditions and select the feedback that best satisfies them.

\medskip
\textbf{[Condition]}\\
1. The feedback \{should/should not\} quote parts of the essay that are relevant to evaluating the given traits (Specificity).\\
2. The feedback \{should/should not\} include actionable revision suggestions for improving the essay (Helpfulness).\\
3. The feedback \{should/should not\} align with the score descriptions in the rubric (Validity).\\
\textbf{(end of [Condition])}\\

\textcolor{BrickRed}{\textbf{[Examples]\\}
Three examples of essays, scores, feedback candidates, and selected feedback\\
\textbf{(end of [Examples])}}

\medskip
\textbf{[Essay]}\\
\{essay text\}\\
\textbf{(end of [Essay])}

\medskip
\textbf{[Scores]}\\
{Narrativity: 3}\\
{Language: 2}\\
(\ldots)\\
\textbf{(end of [Scores])}

\medskip
\textbf{[Rubric descriptions]}\\
\textbf{[Trait]}\\
Narrativity\\
\textbf{(end of [Trait])}\\
The following is a rubric description in terms of the ``Narrativity'' trait.\\
Score 3: The response is interesting. Appropriate use of transition and \ldots\\
\textbf{[Trait]}\\
Language\\
(\ldots)\\
\textbf{(end of [Rubric descriptions])}

\medskip
\textbf{[Feedback Candidates]}

\medskip
``feedback 1'': ``The essay is \ldots''\\
``feedback 2'': ``The essay is \ldots''\\
``feedback 3'': ``The essay is \ldots''\\
``feedback 4'': ``The essay is \ldots''\\
``feedback 5'': ``The essay is \ldots''\\
``feedback 6'': ``The essay is \ldots''\\
``feedback 7'': ``The essay is \ldots''\\
``feedback 8'': ``The essay is \ldots''

\medskip
\textbf{(end of [Feedback Candidates])}

\medskip
\textcolor{Brown}{\textbf{Think step by step.}}\\Pick only one feedback without additional explanation:\\
ex) feedback 3

\end{tcolorbox}
\caption{Prompt template for filtering essay feedback by using LLM-as-a-judge.}
\label{gpt_filtering_prompt}
\end{figure*}

\section{Dataset Details}
\subsection{Datasets for Helpfulness Evaluator} \label{helpfulness_dataset_detail}





The following three sources serve as the primary basis for building chosen–rejected pairs of the datasets for training the helpfulness evaluator:

\begin{itemize}
    \item \textbf{RECIPE4U \cite{recipe4u}:} A dataset collected from university-level English learners who iteratively revised their essays based on LLM-generated feedback. In our setting, feedback that students accepted and used for revision is labeled as chosen, whereas feedback that was not adopted is labeled as rejected.
    \item \textbf{FEAT \cite{feat}:} A dataset in which students read English passages and wrote open-ended responses to comprehension questions. Human annotators ranked feedback—generated by both humans and LLMs—based on its accuracy and helpfulness in improving the student’s answer, and we use these rankings to construct chosen–rejected pairs.
    \item \textbf{ASAP++ \cite{mathias2018++}:} We use 45 human-written feedback instances from the ASAP++ dataset. Inspired by \citet{behzad2024leaf}, we prompt GPT-5.1 to generate revised versions of the human-written feedback, treating the revised feedback as the \textit{chosen} example and the original human-written feedback as the \textit{rejected} example.
\end{itemize}

\subsection{Details of ASAP++ Dataset}
Table \ref{table_asap_detail} summarizes the ASAP++ dataset, including essay characteristics, evaluated traits, and score ranges for each prompt.


\begin{table*}[h]
\centering
\small
\setlength{\tabcolsep}{6pt}
\resizebox{\textwidth}{!}{%
\begin{tabular}{|c|c|c|c|c|c|c|c|c|c|}
\toprule
\textbf{Prompt} &
\textbf{\# of Essays} &
\textbf{Average Length} &
\textbf{Essay Type} &
\textbf{Grade Level} &
\textbf{Traits} &
\multicolumn{2}{c}{\textbf{Score Range}} \\
\cmidrule(lr){7-8}
 & & & & & & \textbf{Overall} & \textbf{Trait} \\
\midrule
P1 & 1,783 & 350 & Argumentative & 8  & Over, Cont, WC, Org, SF, Conv & 2 - 12 & 1 - 6 \\
P2 & 1,800 & 350 & Argumentative & 10 & Over, Cont, WC, Org, SF, Conv & 1 - 6  & 1 - 6 \\
P3 & 1,726 & 150 & Source-Dependent & 10 & Over, Cont, PA, Nar, Lan     & 0 - 3  & 0 - 3 \\
P4 & 1,772 & 150 & Source-Dependent & 10 & Over, Cont, PA, Nar, Lan     & 0 - 3  & 0 - 3 \\
P5 & 1,805 & 150 & Source-Dependent & 8  & Over, Cont, PA, Nar, Lan     & 0 - 4  & 0 - 4 \\
P6 & 1,800 & 150 & Source-Dependent & 10 & Over, Cont, PA, Nar, Lan     & 0 - 4  & 0 - 4 \\
\bottomrule
\end{tabular}
}
\caption{Statistics of the ASAP++ dataset. Traits include Overall (Over), Content (Cont), Word Choice (WC), Organization (Org), Sentence Fluency (SF), Conventions (Conv), Prompt Adherence (PA), Narrativity (Nar), and Language (Lang).}
\label{table_asap_detail}
\end{table*}

\section{Human Evaluation Details}
We conducted various human evaluations with three teacher annotators holding master's degrees in English education.

\subsection{Pairwise Comparison of Essay Feedback Quality across Specificity, Helpfulness, and Validity} \label{detail_of_pairwise_comparison}
Figure \ref{dimension_definition_for_expert} illustrates the definitions and descriptive criteria for the three feedback evaluation dimensions. Prior to the main annotation, the three teacher annotators completed two training rounds with 10 practice pairs per round to establish a shared mental model of the evaluation criteria. Subsequently, we measured inter-rater reliability using Fleiss’ Kappa on a subset of 30 pairs, observing agreement above 0.85, which indicates very high consistency. After verifying the consistency, the annotators labeled 150 pairs per dimension—\textit{specificity}, \textit{validity}, and \textit{helpfulness}—resulting in a total of 450 annotated pairs across all dimensions.

\begin{figure}[]
\centering
\begin{tcolorbox}[
  colback=gray!5,
  colframe=black,
  fonttitle=\bfseries,
]
\small

\medskip
\textbf{Specificity}: Evaluates how concretely the feedback reflects the content of the essay.\\
- To what extent is the essay content explicitly referenced in the feedback?\\
- Is the essay content referenced evenly across the sentences of the feedback?\\
- Does the feedback refer to multiple parts of the essay in a balanced manner?\\

\medskip
\textbf{Helpfulness}: Evaluates the extent to which the feedback supports improvement of the learner’s essay.\\
- Does the feedback identify aspects of the essay that need improvement?\\
- Does the feedback provide sufficient information to help the student improve the essay?\\

\medskip
\textbf{Validity}: Evaluates how well the feedback reflects the level of the essay score based on the rubric criteria.\\
- To what extent does the feedback reflect the rubric criteria corresponding to the essay score?\\
- How accurately does the feedback use expressions from the rubric criteria corresponding to the essay score?

\end{tcolorbox}
\caption{Definitions and detailed descriptions of feedback evaluation dimensions.}
\label{dimension_definition_for_expert}
\end{figure}

\subsection{Human-evaluation of Essay Feedback Filtered by FeedEval}
Initially, the three teacher annotators completed two training sessions, each consisting of 10 practice essay feedback samples, to establish a shared understanding of the feedback evaluation criteria. Inter-rater reliability was then assessed using the intra-class correlation coefficient (ICC) on 30 essay feedback samples, indicating good agreement (D1=0.78, D2=0.72, D3=0.75).

\subsection{Human-evaluation of Essay Revised by Small LLMs}
The three teacher annotators were asked to select the better revised essay from each pair of revision outcomes. They again completed two training sessions with 10 essay pairs per session to calibrate their judgments, achieving high inter-rater consistency with a Fleiss’ Kappa of 0.81 on 30 essay pairs.

\section{Implementation Details} \label{implementation_detail}
Details of the configurations for training essay assessment LLMs and implementing essay revision LLMs are provided below:

\subsection{FeedEval LLMs}
We fine-tune Llama3-3B-Instruct\footnote{\url{https://huggingface.co/meta-llama/Llama-3.2-3B-Instruct}}, Qwen2-3B-Instruct\footnote{\url{https://huggingface.co/Qwen/Qwen2.5-3B-Instruct}}, Phi-3-Mini-Instruct\footnote{\url{https://huggingface.co/microsoft/Phi-3.5-mini-instruct}}, and Gemma3-Instruct\footnote{\url{https://huggingface.co/google/gemma-3-4b-it}} models with DeepSpeed Stage-2 \cite{rasley2020deepspeed} on eight NVIDIA A100 (80GB) GPUs. Training is conducted for 5 epochs (100 steps) using AdamW with a batch size of 4, an initial learning rate of 1e-5. The margin hyperparameter for training the evaluator of specificity and helpfulness is set to 0.5. All other hyperparameters follow default settings.

\subsection{Essay Assessment LLMs}
We fine-tune Llama3-8B-Instruct\footnote{\url{https://huggingface.co/meta-llama/Llama-3.1-8B-Instruct}} and Qwen3-8B\footnote{\url{https://huggingface.co/Qwen/Qwen3-8B}} models using LoRA \cite{hu2022lora} with DeepSpeed Stage-2 \cite{rasley2020deepspeed} on eight NVIDIA A100 (80GB) GPUs. Training is conducted for 5 epochs (100 steps) using AdamW with a batch size of 4, an initial learning rate of 1e-4, a weight decay of 0.05, and early stopping with a patience of 2. All other hyperparameters follow default settings.

\subsection{Essay Revision LLMs}
For essay revision, we prompt Llama3-1B-Instruct\footnote{\url{https://huggingface.co/meta-llama/Llama-3.2-1B-Instruct}} and Qwen2-1.5B-Instruct\footnote{\url{https://huggingface.co/Qwen/Qwen2.5-1.5B-Instruct}} models to revise essays based on provided feedback. We set the temperature to 0.7 and generate up to 1,000 new tokens.

\section{Specificity Evaluator Trained on Llama-generated SpecEval Dataset} \label{llama_speceval}

We generate a new SpecEval dataset using Llama3-70B \footnote{\url{https://huggingface.co/meta-llama/Meta-Llama-3-70B}}. Then we train Llama3-3B-Instruct on the SpecEval and compare its agreement with expert annotations as well as with the Llama3-3B-Instruct trained on GPT-4o-generated SpecEval. As shown in Table \ref{llama_speceval_comparison}, the evaluators trained on the LLaMA-synthesized data demonstrated comparable human–LLM agreement in terms of accuracy and F1 score to its GPT-4o-based counterpart. Furthermore, both evaluators achieved high agreement scores exceeding 0.8 in both metrics.

\begin{table}[h!]
\centering
\renewcommand{\arraystretch}{0.9}
\resizebox{\columnwidth}{!}{%
\begin{tabular}{|c|cc|cc|cc|}
\toprule
\multirow{2}{*}{\textbf{Model}} 
& \multicolumn{2}{c|}{\textbf{GPT vs. Human}} 
& \multicolumn{2}{c|}{\textbf{Llama vs. Human}} 
& \multicolumn{2}{c|}{\textbf{GPT vs. Llama}} \\ \cmidrule{2-7}
& \textbf{Acc.}  & \textbf{F1}    
& \textbf{Acc.}  & \textbf{F1}    
& \textbf{Acc.}  & \textbf{F1}    \\ \midrule
Llama3-3B-Inst.   & \underline{0.820}           & \underline{0.880}      & \underline{0.800}          & \underline{0.864}          & 0.824 & \underline{0.884} \\
Qwen2-3B-Inst.    & 0.807          & 0.870           & 0.778          & 0.858          & 0.812          & 0.879          \\
Phi-3-Mini     & 0.811          & 0.860           & 0.793 & 0.852 & \underline{0.826}          & 0.863          \\ 
Gemma3-Inst.   & \textbf{0.832} & \textbf{0.893} & \textbf{0.827}          & \textbf{0.881}          & \textbf{0.840}          & \textbf{0.902}          \\ \bottomrule
\end{tabular}%
}
\caption{Agreement of specificity evaluators trained on SpecEval datasets synthesized by Llama3-70B (Llama) or GPT-4o (GPT), compared with human experts and between the two evaluators. All 3B-scale models are fine-tuned. The best results are in bold, and the second-best are \underline{underlined}.}
\label{llama_speceval_comparison}
\end{table}

\section{Analysis of Potential Conflicts Between Evaluation Dimensions}

\begin{table}[H]
\centering
\small
\begin{tabular}{|c|ccc|}
\toprule
\textbf{Dimension} & Specificity & Helpfulness & Validity \\ \midrule
Specificity        & -                & 0.597                & 0.535             \\
Helpfulness        & 0.597                & -                & 0.511             \\
Validity           & 0.535                & 0.511                & -             \\ \bottomrule
\end{tabular}
\caption{Pearson Correlation Coefficients Among Evaluation Dimensions}
\label{dimension_correlation}
\end{table}

To evaluate potential conflicts among the three evaluation dimensions—\textit{specificity}, \textit{helpfulness}, and \textit{validity}—the Pearson correlation coefficients between the scores generated by each dimension-specific evaluation model (Llama3-3B-Instruct) were analyzed. As presented in Table \ref{dimension_correlation}, the results indicate moderate positive correlations across all pairs (0.40 < \textit{r} < 0.60). This suggests that the dimension scores exhibit some linear association and are not in opposing directions, while still capturing distinguishable aspects.

\section{Impact of FeedEval on ASAP-SAS Dataset} \label{feedeval_on_sas}
\subsection{ASAP-SAS Dataset}
To examine generalizability, we additionally utilize the ASAP-SAS dataset \cite{asap_sas}\footnote{\url{https://www.kaggle.com/competitions/asap-sas}}
, which contains open-ended student responses across multiple subjects. Since no publicly available dataset other than the ASAP++ provides multi-trait essay scores directly linked to scoring rubrics, we adopt ASAP-SAS as a practical alternative, despite it offering only overall quality labels. We select five prompts related to essay writing, and use this dataset as a complementary benchmark to evaluate the generalizability of FeedEval in essay assessment settings.

\begin{table}[htbp]
\centering
\small
\setlength{\tabcolsep}{6pt}
\begin{tabular}{|>{\centering\arraybackslash}m{1.6cm}|c|c|c|}
    \toprule
    \textbf{Dataset} & \textbf{Prompt} & \textbf{\# Essays} & \textbf{Score Range} \\
    \midrule
    \multirow{5}{*}{\textbf{ASAP-SAS}} &
    3 & 2214 & \multirow{5}{*}{0 - 3} \\ 
    & 4 & 1952 & \\ 
    & 7 & 2398 & \\ 
    & 8 & 2398 & \\ 
    & 9 & 2397 & \\ 
    \bottomrule
\end{tabular}
\caption{Statistics of the ASAP-SAS dataset.}
\label{dataset_description}
\end{table}

\subsection{Essay Scoring Performance}

Table \ref{table_sas_scoring} reports essay scoring performance of models trained on the ASAP-SAS dataset to jointly generate scores and feedback. Consistent with ASAP++, models trained on high-quality feedback filtered by FeedEval outperform those trained on low-quality feedback across all prompts. In contrast, feedback labeled as high-quality by GPT-5.1 does not consistently yield better performance than its low-quality counterparts. Moreover, models trained with FeedEval-selected high-quality feedback consistently surpass those trained with GPT-5.1-selected feedback. These results demonstrate that FeedEval’s feedback quality assessment generalizes beyond ASAP++ to other essay datasets.

\begin{table*}[]
\centering
\resizebox{\textwidth}{!}{%
\begin{tabular}{|ccc|ccccc|c|}
\toprule
 & & &
  \multicolumn{5}{c|}{\textbf{Prompt}} & \\ \midrule
\multicolumn{1}{|c|}{\textbf{LLM}} &
  \multicolumn{1}{c|}{\textbf{Assessment}} &
  \textbf{Feedback Quality} &
  \textbf{3} &
  \textbf{4} &
  \textbf{7} &
  \textbf{8} &
  \textbf{9} &
  \textbf{Avg $\uparrow$ (SD $\downarrow$)} \\ \midrule

\multicolumn{1}{|c|}{} &
  \multicolumn{1}{c|}{Score Only} &
  \xmark &
  0.556 & 0.551 & 0.534 & 0.527 & 0.634 &
  0.560 ($\pm$0.037) \\ \cmidrule{2-9}

\multicolumn{1}{|c|}{} &
  \multicolumn{1}{c|}{\multirow{2}{*}{\centering Score + Feedback}} &
  Low Quality (GPT-5.1) &
  0.589 & 0.602 & 0.498 & 0.551 & 0.695 &
  0.587 ($\pm$0.027) \\
\multicolumn{1}{|c|}{} &
  \multicolumn{1}{c|}{} &
  High Quality (GPT-5.1) &
  \underline{0.641} & 0.597 & \underline{0.539} & 0.549 & \underline{0.709} &
  \underline{0.607} ($\pm$0.029) \\ \cmidrule{2-9}

\multicolumn{1}{|c|}{Llama3-8B-Inst.} &
  \multicolumn{2}{c|}{\textbf{Improvement (High Quality - Low Quality)}} &
  \cellcolor{blue4}+8.83\% &
  \cellcolor{red2}-0.83\% &
  \cellcolor{blue4}+8.23\% &
  \cellcolor{red2}-0.36\% &
  \cellcolor{blue2}+2.01\% &
  \cellcolor{blue2}+3.41\% \\ \cmidrule{2-9}

\multicolumn{1}{|c|}{} &
  \multicolumn{1}{c|}{\multirow{2}{*}{\centering Score + Feedback}} &
  Low Quality (FeedEval) &
  0.617 & \underline{0.604} & 0.521 & \underline{0.563} & 0.704 &
  0.602 ($\pm$0.030) \\
\multicolumn{1}{|c|}{} &
  \multicolumn{1}{c|}{} &
  High Quality (FeedEval) &
  \textbf{0.661} & \textbf{0.610} & \textbf{0.551} & \textbf{0.595} & \textbf{0.729} &
  \textbf{0.629} ($\pm$0.035) \\ \cmidrule{2-9}

\multicolumn{1}{|c|}{} &
  \multicolumn{2}{c|}{\textbf{Improvement (High Quality - Low Quality)}} &
  \cellcolor{blue4}+7.13\% &
  \cellcolor{blue2}+0.99\% &
  \cellcolor{blue4}+5.76\% &
  \cellcolor{blue4}+5.68\% &
  \cellcolor{blue2}+3.55\% &
  \cellcolor{blue2}+4.55\% \\ \midrule

\multicolumn{1}{|c|}{} &
  \multicolumn{1}{c|}{Score Only} &
  \xmark &
  0.570 & \underline{0.585} & 0.483 & 0.542 & 0.691 &
  0.574 ($\pm$0.027) \\ \cmidrule{2-9}

\multicolumn{1}{|c|}{} &
  \multicolumn{1}{c|}{\multirow{2}{*}{\centering Score + Feedback}} &
  Low Quality (GPT-5.1) &
  0.565 & 0.581 & 0.454 & 0.537 & 0.681 &
  0.564 ($\pm$0.031) \\
\multicolumn{1}{|c|}{} &
  \multicolumn{1}{c|}{} &
  High Quality (GPT-5.1) &
  0.580 & 0.577 & 0.479 & 0.549 & 0.673 &
  0.572 ($\pm$0.037) \\ \cmidrule{2-9}

\multicolumn{1}{|c|}{Qwen3-8B} &
  \multicolumn{2}{c|}{\textbf{Improvement (High Quality - Low Quality)}} &
  \cellcolor{blue2}+2.65\% &
  \cellcolor{red2}-0.69\% &
  \cellcolor{blue4}+5.51\% &
  \cellcolor{blue2}+2.23\% &
  \cellcolor{red2}-1.17\% &
  \cellcolor{blue2}+1.42\% \\ \cmidrule{2-9}

\multicolumn{1}{|c|}{} &
  \multicolumn{1}{c|}{\multirow{2}{*}{\centering Score + Feedback}} &
  Low Quality (FeedEval) &
  \underline{0.582} & 0.577 & \underline{0.491} & \underline{0.553} & \underline{0.706} &
  \underline{0.582} ($\pm$0.035) \\
\multicolumn{1}{|c|}{} &
  \multicolumn{1}{c|}{} &
  High Quality (FeedEval) &
  \textbf{0.606} & \textbf{0.617} & \textbf{0.521} & \textbf{0.563} & \textbf{0.726} &
  \textbf{0.607} ($\pm$0.030) \\ \cmidrule{2-9}

\multicolumn{1}{|c|}{} &
  \multicolumn{2}{c|}{\textbf{Improvement (High Quality - Low Quality)}} &
  \cellcolor{blue2}+4.12\% &
  \cellcolor{blue4}+6.93\% &
  \cellcolor{blue4}+6.11\% &
  \cellcolor{blue2}+1.81\% &
  \cellcolor{blue2}+2.83\% &
  \cellcolor{blue2}+4.26\% \\ \bottomrule

\end{tabular}%
}
\caption{Average essay scoring performance for each prompt on the ASAP-SAS dataset. We report five-fold averaged results with standard deviations (SD). The best performances are shown in \textbf{bold}, and the second-best are \underline{underlined} for each LLM backbone.}
\label{table_sas_scoring}
\end{table*}

\subsection{Essay Improvement after Revision}
\begin{figure}[htbp]
    \centering
    \includegraphics[width=\linewidth]{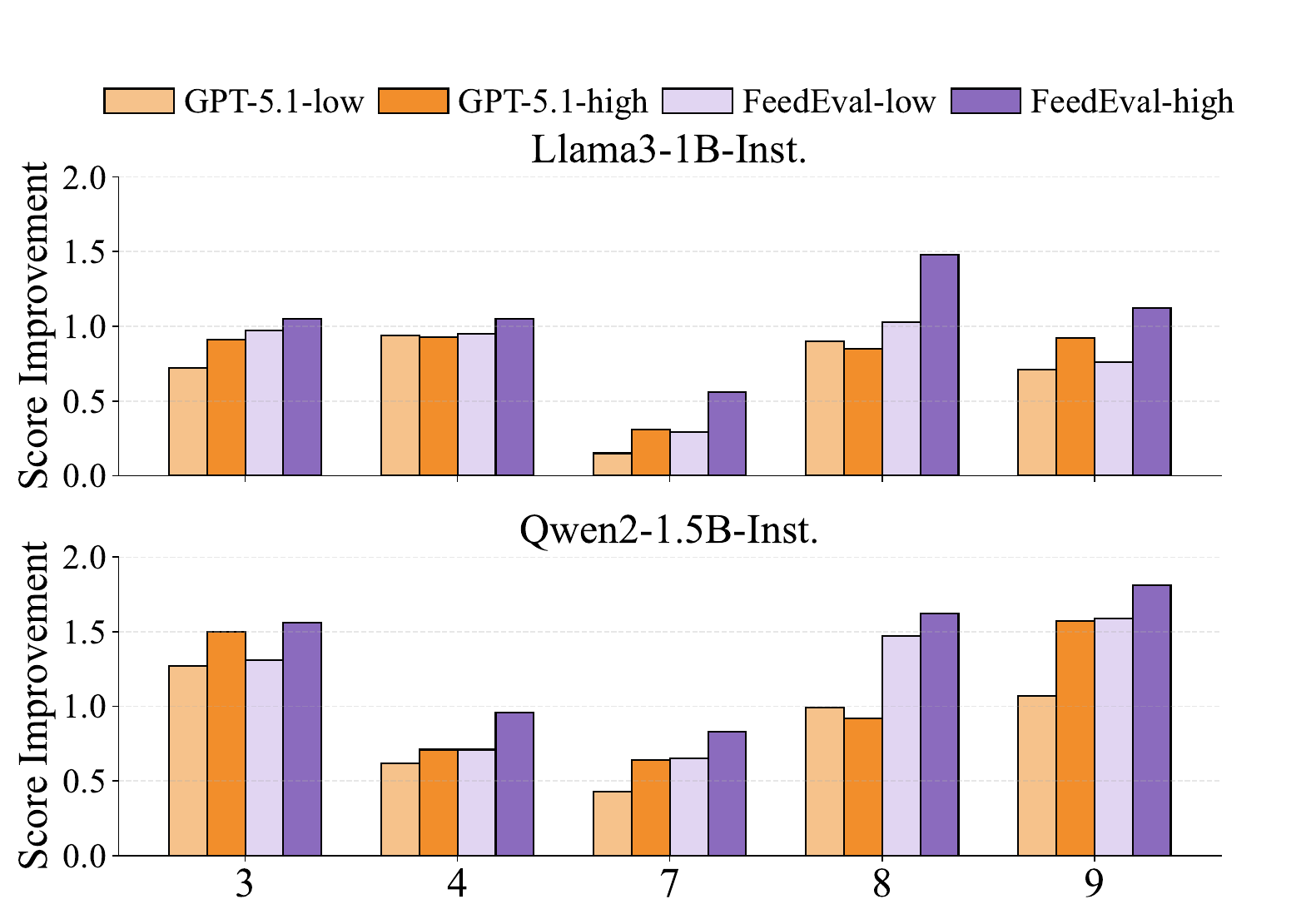}
    \caption{Average essay score improvement across prompts on ASAP-SAS after revisions guided by feedback of high- and low-quality identified by FeedEval and GPT-5.1.}
    \label{fig_sas_revision}
\end{figure}
Figure \ref{fig_sas_revision} shows average prompt-level score improvements when Llama3-1B-Instruct and Qwen2-1.5B-Instruct revise essays using high- or low-quality feedback identified by FeedEval and GPT-5.1. Feedback filtered as high-quality by FeedEval consistently yields larger revision gains across traits, whereas GPT-5.1-filtered feedback does not show consistent improvements over its low-quality counterpart, indicating limited discriminative ability. Moreover, FeedEval-selected high-quality feedback leads to greater gains than GPT-5.1–selected feedback. Overall, these results demonstrate that FeedEval reliably identifies high-quality feedback that enables more effective essay revisions on ASAP-SAS, supporting its generalizability.

\section{Essay Assessment Performance with Reversely Constructed Labels} \label{reverse_performance}
Table~\ref{table_reverse} reports the essay scoring performance of Llama3-8B-Instruct and Qwen3-8B trained on high- and low-quality feedback datasets filtered by FeedEval, where models are trained to generate feedback before predicting scores. While training on high-quality feedback still yields consistent improvements over low-quality feedback across traits, overall scoring performance is lower than that of models trained to predict scores first, indicating that the order of score and feedback generation affects assessment accuracy \cite{do2025radme}. This suggests that LLMs benefit more from post-thinking mechanisms than from pre-thinking mechanisms in essay scoring tasks \cite{chen2025distilling}.

\begin{table*}[]
\centering
\resizebox{\textwidth}{!}{%
\begin{tabular}{|ccc|ccccccccc|c|}
\toprule
\multicolumn{3}{|c|}{} &
  \multicolumn{9}{c|}{\textbf{Traits (Prediction Order: $\leftarrow$)}} &
   \\ \midrule
\multicolumn{1}{|c|}{\textbf{LLM}} &
  \multicolumn{1}{c|}{\textbf{Assessment}} &
  \textbf{Feedback Quality} &
  \textbf{Over} &
  \textbf{Cont} &
  \textbf{PA} &
  \textbf{Lang} &
  \textbf{Nar} &
  \textbf{Org} &
  \textbf{Conv} &
  \textbf{WC} &
  \textbf{SF} &
  \textbf{Avg$\uparrow$ (SD$\downarrow$)} \\ \midrule
\multicolumn{1}{|c|}{\multirow{3}{*}{Llama3-8B-Inst.}} &
  \multicolumn{1}{c|}{\multirow{2}{*}{Score+Feedback}} &
  Low Quality &
  0.455 & 0.504 & 0.561 & 0.534 & 0.579 & 0.453 & 0.443 & 0.496 & 0.502 &
  0.503 ($\pm$0.042) \\
\multicolumn{1}{|c|}{} &
  \multicolumn{1}{c|}{} &
  High Quality &
  0.464 & 0.508 & 0.590 & 0.538 & 0.600 & 0.470 & 0.461 & 0.530 & 0.527 &
  0.521 ($\pm$0.032) \\ \cmidrule{2-13}
\multicolumn{1}{|c|}{} &
  \multicolumn{2}{c|}{\textbf{Improvement (High Quality - Low Quality)}} &
  \cellcolor{blue2}+1.98\% & \cellcolor{blue2}+0.79\% & \cellcolor{blue4}+5.17\% & \cellcolor{blue2}+0.75\% & \cellcolor{blue2}+3.63\% & \cellcolor{blue2}+3.75\% & \cellcolor{blue2}+4.06\% & \cellcolor{blue4}+6.85\% & \cellcolor{blue2}+4.98\% &
  \cellcolor{blue2}+3.56\% \\ \midrule

\multicolumn{1}{|c|}{\multirow{3}{*}{Qwen3-8B}} &
  \multicolumn{1}{c|}{\multirow{2}{*}{Score+Feedback}} &
  Low Quality &
  0.637 & 0.607 & 0.613 & 0.614 & 0.624 & 0.664 & 0.691 & 0.646 & 0.684 &
  0.642 ($\pm$0.024) \\
\multicolumn{1}{|c|}{} &
  \multicolumn{1}{c|}{} &
  High Quality &
  0.648 & 0.626 & 0.618 & 0.662 & 0.626 & 0.679 & 0.695 & 0.650 & 0.698 &
  0.656 ($\pm$0.026) \\ \cmidrule{2-13}
\multicolumn{1}{|c|}{} &
  \multicolumn{2}{c|}{\textbf{Improvement (High Quality - Low Quality)}} &
  \cellcolor{blue2}+1.73\% & \cellcolor{blue2}+3.13\% & \cellcolor{blue2}+0.82\% & \cellcolor{blue4}+7.82\% & \cellcolor{blue2}+0.32\% & \cellcolor{blue2}+2.26\% & \cellcolor{blue2}+0.58\% & \cellcolor{blue2}+0.62\% & \cellcolor{blue2}+2.05\% &
  \cellcolor{blue2}+2.11\% \\ \bottomrule
\end{tabular}%
}
\caption{Average essay scoring performance across all prompts for each trait on the ASAP++ dataset. Traits are predicted from right to left ($\leftarrow$). We report five-fold averaged results with standard deviations (SD).}
\label{table_reverse}
\end{table*}

\section{Essay Scoring Performance under Different FeedEval Dimension Configurations} \label{essay_scoring_dimension_config}
Table \ref{table_ablation} reports the essay scoring performance of Qwen3-8B trained on high-quality feedback filtered by FeedEval under different dimension configurations. For multi-dimensional configuration, feedback is selected based on the average FeedEval score computed over the corresponding subset of dimensions.

\begin{table*}[]
\centering
\resizebox{\textwidth}{!}{%
\begin{tabular}{|c|c|ccccccccc|c|}
\toprule
    &                         & \multicolumn{9}{c|}{\textbf{Traits (Prediction Order: $\leftarrow$)}}              &               \\ \midrule
\textbf{\# of Dimensions} &
  \textbf{Dimension} &
  \textbf{Over} &
  \textbf{Cont} &
  \textbf{PA} &
  \textbf{Lang} &
  \textbf{Nar} &
  \textbf{Org} &
  \textbf{Conv} &
  \textbf{WC} &
  \textbf{SF} &
  \textbf{Avg$\uparrow$ (SD$\downarrow$)} \\ \midrule
    & Specificity             & 0.657 & 0.689 & 0.700 & \underline{0.678} & 0.712 & 0.669 & 0.688 & 0.672 & 0.681 & 0.683 ($\pm$0.021) \\
One & Helpfulness             & 0.652 & 0.686 & 0.701 & \textbf{0.683} & \underline{0.717} & 0.668 & 0.681 & 0.671 & 0.685 & 0.683 ($\pm$0.024) \\
    & Validity                & 0.661 & 0.685 & 0.703 & 0.672 & 0.707 & \underline{0.673} & 0.683 & 0.684 & 0.682 & 0.683 ($\pm$0.025) \\ \midrule
    & Specificity+Helpfulness & 0.654 & 0.691 & 0.698 & 0.673 & 0.709 & \textbf{0.675} & 0.686 & 0.681 & 0.685 & 0.684 ($\pm$0.020) \\
Two & Specificity+Validity    & \underline{0.662} & \underline{0.692} & 0.701 & \underline{0.678} & 0.708 & 0.672 & 0.691 & 0.683 & 0.696 & 0.687 ($\pm$0.025) \\
    & Helpfulness+Validity    & \textbf{0.664} & 0.69  & \underline{0.706} & 0.677 & 0.71  & 0.67  & \textbf{0.697} & \underline{0.685} & \textbf{0.703} & \underline{0.689} ($\pm$0.029) \\ \midrule
Three &
  \multicolumn{1}{l|}{Specificity+Helpfulness+Validity} &
  0.661 &
  \textbf{0.699} &
  \textbf{0.709} &
  \textbf{0.683} &
  \textbf{0.719} &
  \underline{0.673} &
  \underline{0.694} &
  \textbf{0.688} &
  \underline{0.698} &
  \textbf{0.692} ($\pm$0.020) \\ \bottomrule
\end{tabular}%
}
\caption{Average essay scoring performance on the ASAP++ dataset using high-quality feedback filtered by different FeedEval dimension configurations, averaged across all prompts for each trait. Traits are predicted from right to left ($\leftarrow$). We report five-fold averaged results with standard deviations (SD).  The best performances are shown in \textbf{bold}, and the second-best are \underline{underlined}.}
\label{table_ablation}
\end{table*}

\section{Additional Experiments}
\subsection{Essay Scoring Performance averaged across the traits for each prompt}
Table \ref{table_prompt_wise} reports the essay scoring performance of Llama3-8B-Instruct and Qwen3-8B trained on different feedback quality, averaged across traits for each prompt. Models trained with high-quality feedback filtered by FeedEval consistently outperform those trained with low-quality feedback across all prompts, demonstrating the effectiveness of high-quality feedback as supervision for essay scoring. In contrast, models trained on feedback filtered as high-quality by GPT-5.1 do not yield consistent performance gains over their low-quality counterparts. Moreover, models trained on FeedEval-selected high-quality feedback consistently outperform those trained on GPT-5.1-selected high-quality feedback, highlighting FeedEval’s superior ability to assess and filter pedagogically useful feedback.


\begin{table*}[]
\centering
\resizebox{\textwidth}{!}{%
\begin{tabular}{|ccc|ccccccc|}
\toprule
 &
   &
   &
  \multicolumn{7}{c|}{\textbf{Prompt}} \\ \midrule
\multicolumn{1}{|c|}{\textbf{LLM}} &
  \multicolumn{1}{c|}{\textbf{Assessment}} &
  \textbf{Feedback Quality} &
  \textbf{1} &
  \textbf{2} &
  \textbf{3} &
  \textbf{4} &
  \textbf{5} &
  \multicolumn{1}{c|}{\textbf{6}} &
  \textbf{Avg} \\ \midrule

\multicolumn{1}{|c|}{} &
  \multicolumn{1}{c|}{Score Only} &
  \xmark &
  0.430 & 0.519 & \textbf{0.600} & 0.659 & \textbf{0.567} &
  \multicolumn{1}{c|}{0.514} &
  0.548 (0.042) \\ \cmidrule{2-10}

\multicolumn{1}{|c|}{} &
  \multicolumn{1}{c|}{\multirow{2}{*}{\centering Score + Feedback}} &
  Low Quality (GPT-5.1) &
  0.455 & 0.526 & 0.555 & \underline{0.675} & 0.531 &
  \multicolumn{1}{c|}{\underline{0.585}} &
  0.555 (0.041) \\
\multicolumn{1}{|c|}{} &
  \multicolumn{1}{c|}{} &
  High Quality (GPT-5.1) &
  0.443 & 0.517 & 0.567 & 0.668 & 0.544 &
  \multicolumn{1}{c|}{0.584} &
  0.554 (0.043) \\ \cmidrule{2-10}

\multicolumn{1}{|c|}{Llama3-8B-Inst.} &
  \multicolumn{2}{c|}{\textbf{Improvement (High Quality - Low Quality)}} &
  \cellcolor{red2}-2.64\% & \cellcolor{red2}-1.71\% & \cellcolor{blue2}+2.16\% & \cellcolor{red2}-1.04\% & \cellcolor{blue2}+2.45\% &
  \multicolumn{1}{c|}{\cellcolor{red2}-0.17\%} &
  \cellcolor{red2}-0.12\% \\ \cmidrule{2-10}

\multicolumn{1}{|c|}{} &
  \multicolumn{1}{c|}{\multirow{2}{*}{\centering Score + Feedback}} &
  Low Quality (FeedEval) &
  \underline{0.479} & \underline{0.537} & 0.553 & 0.674 & 0.540 &
  \multicolumn{1}{c|}{0.564} &
  \underline{0.558} (0.047) \\
\multicolumn{1}{|c|}{} &
  \multicolumn{1}{c|}{} &
  High Quality (FeedEval) &
  \textbf{0.492} & \textbf{0.544} & \underline{0.588} & \textbf{0.689} & \underline{0.547} &
  \multicolumn{1}{c|}{\textbf{0.587}} &
  \textbf{0.575} (0.040) \\ \cmidrule{2-10}

\multicolumn{1}{|c|}{} &
  \multicolumn{2}{c|}{\textbf{Improvement (High Quality - Low Quality)}} &
  \cellcolor{blue2}+2.71\% & \cellcolor{blue2}+1.30\% & \cellcolor{blue4}+6.33\% & \cellcolor{blue2}+2.23\% & \cellcolor{blue2}+1.30\% &
  \multicolumn{1}{c|}{\cellcolor{blue2}+4.08\%} &
  \cellcolor{blue2}+2.99\% \\ \midrule

\multicolumn{1}{|c|}{} &
  \multicolumn{1}{c|}{Score Only} &
  \xmark &
  \textbf{0.687} & \underline{0.662} & 0.695 & 0.747 & \underline{0.671} &
  \multicolumn{1}{c|}{0.685} &
  \underline{0.690} (0.025) \\ \cmidrule{2-10}

\multicolumn{1}{|c|}{} &
  \multicolumn{1}{c|}{\multirow{2}{*}{\centering Score + Feedback}} &
  Low Quality (GPT-5.1) &
  0.618 & 0.648 & 0.695 & \underline{0.756} & 0.663 &
  \multicolumn{1}{c|}{\underline{0.702}} &
  0.680 (0.029) \\
\multicolumn{1}{|c|}{} &
  \multicolumn{1}{c|}{} &
  High Quality (GPT-5.1) &
  0.631 & 0.645 & \underline{0.701} & 0.753 & \underline{0.671} &
  \multicolumn{1}{c|}{0.689} &
  0.682 (0.034) \\ \cmidrule{2-10}

\multicolumn{1}{|c|}{Qwen3-8B} &
  \multicolumn{2}{c|}{\textbf{Improvement (High Quality - Low Quality)}} &
  \cellcolor{blue2}+2.10\% & \cellcolor{red2}-0.46\% & \cellcolor{blue2}+0.86\% & \cellcolor{red2}-0.40\% & \cellcolor{blue2}+1.21\% &
  \multicolumn{1}{c|}{\cellcolor{red2}-1.85\%} &
  \cellcolor{blue2}+0.20\% \\ \cmidrule{2-10}

\multicolumn{1}{|c|}{} &
  \multicolumn{1}{c|}{\multirow{2}{*}{\centering Score + Feedback}} &
  Low Quality (FeedEval) &
  0.638 & 0.647 & \underline{0.701} & 0.752 & 0.667 &
  \multicolumn{1}{c|}{0.687} &
  0.682 (0.027) \\
\multicolumn{1}{|c|}{} &
  \multicolumn{1}{c|}{} &
  High Quality (FeedEval) &
  \underline{0.639} & \textbf{0.665} & \textbf{0.705} & \textbf{0.765} & \textbf{0.675} &
  \multicolumn{1}{c|}{\textbf{0.711}} &
  \textbf{0.693} (0.021) \\ \cmidrule{2-10}

\multicolumn{1}{|c|}{} &
  \multicolumn{2}{c|}{\textbf{Improvement (High Quality - Low Quality)}} &
  \cellcolor{blue2}+0.16\% & \cellcolor{blue2}+2.78\% & \cellcolor{blue2}+0.57\% & \cellcolor{blue2}+1.73\% & \cellcolor{blue2}+1.20\% &
  \multicolumn{1}{c|}{\cellcolor{blue2}+3.49\%} &
  \cellcolor{blue2}+1.66\% \\ \bottomrule
\end{tabular}%
}
\caption{Average essay scoring performance across all traits for each prompt on the ASAP++ dataset. We report five-fold averaged results with standard deviations (SD). The best performances are shown in \textbf{bold}, and the second-best are \underline{underlined} for each LLM backbone.}
\label{table_prompt_wise}
\end{table*}

\subsection{Alignment of FeedEval Across Model Scales}

To further examine FeedEval’s alignment with human experts in judging essay feedback quality, we compare Llama3-Instruct models of different parameter sizes fine-tuned on dimension-specific datasets. As shown in Table \ref{table_scalability}, alignment with human experts improves as model size increases. Ultimately, we adopt the Llama3-3B-Instruct model as the backbone for FeedEval, as it offers a reasonable balance between strong alignment with human experts and computational efficiency.


\begin{table}[]
\centering
\resizebox{\columnwidth}{!}{%
\begin{tabular}{|c|cc|cc|cc|}
\toprule
\multirow{2}{*}{\textbf{Model}} 
& \multicolumn{2}{c|}{\textbf{Specificity}} 
& \multicolumn{2}{c|}{\textbf{Helpfulness}} 
& \multicolumn{2}{c|}{\textbf{Validity}} \\ \cmidrule{2-7}
& \textbf{Acc.}  & \textbf{F1}    
& \textbf{Acc.}  & \textbf{F1}    
& \textbf{Acc.}  & \textbf{F1}    \\ \midrule
Llama3-Inst. (1B)   & 0.793           & 0.857           & 0.751          & 0.838          & 0.682 & 0.564 \\
Llama3-Inst. (3B)   & \underline{0.820}          & \underline{0.880}           & \underline{0.864}          & \underline{0.912}          & \underline{0.835}          & \underline{0.709}          \\
Llama3-Inst. (8B)    & \textbf{0.829}          & \textbf{0.897}           & \textbf{0.881} & \textbf{0.926} & \textbf{0.858}          & \textbf{0.722}          \\ \bottomrule
\end{tabular}%
}
\caption{Human alignment of FeedEval using Llama3-Instruct across different parameter scales (pairwise Acc./F1). The best performances are shown in \textbf{bold}, and the second-best are \underline{underlined}.}
\label{table_scalability}
\end{table}

\subsection{Error Analysis}
We analyze error cases in which essay scoring LLMs fail to generate outputs in the expected score-feedback JSON format when trained with feedback of varying quality. Such errors are commonly observed when LLMs are required to generate text in a predefined structured format \cite{wang2025verifiable, do2024}. Aggregating errors across all folds, we observe that on the ASAP++ dataset, Llama3-8B-Instruct exhibits a total error rate of 1.6\% across five folds, whereas Qwen3-8B shows a substantially lower error rate of 0.3\%. In contrast, no formatting errors are observed for any model on the ASAP-SAS dataset. Since the primary focus of this study is essay scoring performance rather than error-rate reduction, we exclude these error cases when computing QWK scores.

\subsection{Case Study of FeedEval-filtered Feedback} \label{appendix_case_study_feedback}

Table \ref{table_case_feedback} compares high- and low-quality feedback from three FeedEval dimensions.
Text highlighted in {\sethlcolor{yellow!55}\hl{yellow}} indicates excerpts that directly reference specific parts of the essay (specificity).
Text highlighted in {\sethlcolor{cyan!35}\hl{blue}} represents actionable revision suggestions (helpfulness).
Text highlighted in {\sethlcolor{green!25}\hl{green}} marks content aligned with the rubric description (validity).
Overall, the high-quality feedback covers a broader range of elements related to specificity, helpfulness, and validity than the low-quality feedback.

\begin{table*}[]
\centering
\scriptsize
\resizebox{\textwidth}{!}{%
\begin{tabular}{|c|m{0.85\textwidth}|}
\toprule
\textbf{Trait}                 & \multicolumn{1}{c|}{Sentence Fluency (Score: 3/6), Word Choice (Score: 3/6), Conventions (Score: 3/6), Organization (Score: 4/6), Content (Score: 3/6)} \\ \toprule
\textbf{Essay}                 & In the @LOCATION1 we have the technology of a computer. Some say that the computers are good for the society. I disagree, I believe that it is bad for a few reasons. Some of the reasons are obesity, cramps, more {\sethlcolor{yellow!55}\hl{sexual harrassment}} and even cyber bullying. {\sethlcolor{yellow!55}\hl{First if people don't get off the computers and go out to exercise then it will cause obesity.}} Think of it this way, if you watch a kid that didn't have to go to school his entire life and he started off at @NUM1 pounds. The only thing the kid will want to do is play on the computer and he will gain weight. Next, tip-top-tip-top, that's all you hear when a kid is on the computer, @CAPS1 teens come home from school and go straight to the computer and don't get off about on a school night. These are the kids that are @CAPS1 likely to get bad cramps. they get the cramps form typing to fast, hard and too long. They also can get the cramps from just sitting down for to long then trying to get up but can't because it hurts to much to decide to move anywhere. There's also a lot more of sexual harassment that is going on some kids goin to a chat room to talk to their friends about some great news, but when they get out of the chat room they are all mad and pissed off. The reason is because someone was making fun of them and they didn't like it. This happens maybe close to @PERCENT1 of the time that teens and even parents go into chat rooms online. This is the third reason on why computers have an huge effect on people. The last reason that computers have a huge effect on people is because of cyberbullying. cyberbullying is when someone bullies you but just online instead. Have you realized that children became poor or sick because they {\sethlcolor{yellow!55}\hl{dont}} want to be hurt or killed over something {\sethlcolor{yellow!55}\hl{very stupid}}? Well, I have and {\sethlcolor{yellow!55}\hl{it's not nice}}, and now the god thing is if a kid goes to school online he/she @MONTH1 get cyberbullied because he's not smart enough or if he/she don't have the money to go to school. The cyberbullying normally {\sethlcolor{yellow!55}\hl{occures}} in a chat room online. There is where all of the family will either become broke with no money or food. They @MONTH1 even lose their kid or the house that they live in because {\sethlcolor{yellow!55}\hl{someone wanted to act all big and bad to be popular in school.}} In conclusion, these are only a few ways that computers effects people. {\sethlcolor{yellow!55}\hl{The obesity, cramps, sexual harrassment and cyberbullying are not a joke and it's not fun to get done to you.}} Now it's your turn to tell us what you think about the computers. Do you think computers are good and benefits society or is it bad a {\sethlcolor{yellow!55}\hl{rot little kids brains}} to make their lives horrible for them to live with?                                               \\ \midrule
\makecell[c]{\textbf{High-quality} \\ \textbf{Feedback}} &   \textbf{Sentence Fluency}: The essay contains sentences such as {\sethlcolor{yellow!55}\hl{`First if people don't get off the computers and go out to exercise then it will cause obesity,'}} which {\sethlcolor{green!25}\hl{lacks fluidity}} and clearer punctuation. {\sethlcolor{green!25}\hl{Many sentences feature repetitive patterns and awkward constructions}} like {\sethlcolor{yellow!55}\hl{`The obesity, cramps, sexual harassment and cyberbullying are not a joke and it's not fun to get done to you.'}} {\sethlcolor{cyan!35}\hl{Improvement can be made by varying sentence structures and adding transitional elements for smoother flow.}}

\textbf{Word Choice}: Word choices such as {\sethlcolor{yellow!55}\hl{`it's not nice'}} and {\sethlcolor{yellow!55}\hl{`act all big and bad'}} {\sethlcolor{green!25}\hl{lack specificity and precision, making the language seem mundane.}} Expressions like {\sethlcolor{yellow!55}\hl{`very stupid'}} and {\sethlcolor{yellow!55}\hl{`rot little kids brains'}} {\sethlcolor{green!25}\hl{do not effectively convey the intended seriousness of the issues.}} {\sethlcolor{cyan!35}\hl{Enhancing the vocabulary with more precise and varied language would make the argument more compelling.}}

\textbf{Conventions}: The essay exhibits issues with conventions, notably in {\sethlcolor{green!25}\hl{the lack of proper capitalization, punctuation, and spelling errors}}, such as {\sethlcolor{yellow!55}\hl{`sexual harrassment,'}} {\sethlcolor{yellow!55}\hl{`occures,'}} and {\sethlcolor{yellow!55}\hl{`dont.'}} There are {\sethlcolor{green!25}\hl{frequent grammatical errors that disrupt readability and comprehension.}} {\sethlcolor{cyan!35}\hl{To enhance conventions, careful proofreading and editing should focus on correcting spelling, capitalization, punctuation, and grammatical errors.}}

\textbf{Organization}: {\sethlcolor{green!25}\hl{The essay has a basic organizational structure with an introduction, body, and conclusion}} but does not fully function as a cohesive letter to the editor. Ideas are somewhat scattered throughout the paragraphs, affecting the clear flow of arguments. {\sethlcolor{cyan!35}\hl{Refining the format to more clearly differentiate and develop each argument in separate, well-connected paragraphs and ensuring it adheres to the letter format will enhance organization.}}

\textbf{Content}: The essay {\sethlcolor{green!25}\hl{attempts to cover main ideas such as obesity and cyberbullying but does so in a simplistic and unstructured manner.}} Details provided, like {\sethlcolor{yellow!55}\hl{`someone wanted to act all big and bad to be popular in school,'}} {\sethlcolor{green!25}\hl{lack depth and sometimes drift off-topic.}} {\sethlcolor{cyan!35}\hl{Expanding on each issue with more thoughtful and specific examples would provide a deeper exploration of the problems discussed.}}
\\ \midrule

\makecell[c]{\textbf{Low-quality} \\ \textbf{Feedback}}  &   \textbf{Sentence Fluency}: The sentence fluency score reflects a mix of clear and {\sethlcolor{green!25}\hl{awkward sentences throughout the essay}} such as {\sethlcolor{yellow!55}\hl{`The obesity, cramps, sexual harassment and cyberbullying are not a joke and it's not fun to get done to you.'}} While the writer attempts to use a conversational style, many sentences are fragmented or run-on, making it difficult to follow the argument smoothly. {\sethlcolor{green!25}\hl{The lack of variety in sentence structure further limits the fluid}} reading experience. 

\textbf{Word Choice}: The word choice in the essay is {\sethlcolor{green!25}\hl{basic and repetitive}}, which does not effectively convey a sophisticated understanding of the topic. There are attempts to use descriptive language, but the expression like {\sethlcolor{yellow!55}\hl{`rot little kids brains'}} {\sethlcolor{green!25}\hl{is often imprecise}} or informal, detracting from the persuasive impact. Additionally, some words and phrases are used inaccurately, weakening the overall argument.

\textbf{Conventions}: The essay contains several {\sethlcolor{green!25}\hl{grammatical errors}}, such as incorrect verb forms, {\sethlcolor{green!25}\hl{punctuation mistakes, and capitalization issues, which disrupt the readability}}. These errors suggest a lack of careful proofreading and editing. While some sentences are free of errors, the frequency of mistakes throughout the essay impacts the clarity and professionalism of the writing. {\sethlcolor{cyan!35}\hl{To enhance conventions, the writer should focus on correcting spelling, capitalization, and grammatical errors.}}

\textbf{Organization}: {\sethlcolor{green!25}\hl{The organization of the essay shows a logical progression of ideas with a clear introduction, body, and conclusion.}} The writer presents several reasons supporting their opinion against the use of computers, although transitions between ideas could be smoother. Despite some structural weaknesses, the overall organization helps convey the writer's main points effectively.

\textbf{Content}: The content of the essay includes several arguments against the use of computers, such as obesity and cyberbullying, {\sethlcolor{green!25}\hl{but lacks depth and supporting evidence}}. The points are presented {\sethlcolor{cyan!35}\hl{without substantial elaboration or examples that would strengthen the argument}}. Consequently, the essay provides a basic exploration of the topic without fully persuading the reader.\\ \midrule

\makecell[c]{\textbf{Rubric} \\ \textbf{Description}}    & \textbf{Sentence Fluency (Score 3)}: The writing tends to be mechanical {\sethlcolor{green!25}\hl{rather than fluid.}} Occasional {\sethlcolor{green!25}\hl{awkward
constructions may force the reader to slow down or reread}}. The writing is characterized by

-some passages that invite fluid oral reading; however, others do not.

-{\sethlcolor{green!25}\hl{some variety in sentence structure, length, and beginnings, although the writer
falls into repetitive sentence patterns}}.

\textbf{Word Choice (Score 3)}: {\sethlcolor{green!25}\hl{Language lacks precision and variety, or may be inappropriate to audience and
purpose in places.}} The writer does not employ a variety of words, producing a sort of `generic'
paper filled with familiar words and phrases. The writing is characterized by

-words that work, but that rarely capture the reader’s interest.

-{\sethlcolor{green!25}\hl{expression that seems mundane and general}}; {\sethlcolor{green!25}\hl{slang}}, if used, {\sethlcolor{green!25}\hl{does not seem purposeful}} and is not effective.

-attempts at colorful language that seem overdone or forced.

\textbf{Conventions (Score 3)}: {\sethlcolor{green!25}\hl{The writing demonstrates limited control of standard writing conventions (e.g.,
punctuation, spelling, capitalization, grammar and usage)}}. {\sethlcolor{green!25}\hl{Errors begin to impede readability.}}
The writing is characterized by

-some control over basic conventions; the text may be too simple or too short to
reveal mastery.

-end-of-sentence punctuation that is usually correct; however, internal punctuation
contains frequent errors.

-{\sethlcolor{green!25}\hl{spelling errors that distract the reader}}; misspelling of common words occurs.

-{\sethlcolor{green!25}\hl{errors in grammar and usage that do not block meaning but do distract the
reader.}}

\textbf{Organization (Score 4)}: The essay shows satisfactory organization. It contains a basic introduction, body and conclusion.

\textbf{Content (Score 3)}: {\sethlcolor{green!25}\hl{The reader can understand the main ideas}}, although they may be overly broad or
simplistic, and the results may not be effective. {\sethlcolor{green!25}\hl{Supporting detail is often limited, insubstantial,
overly general, or occasionally slightly off-topic.}} The writing is characterized by

-{\sethlcolor{green!25}\hl{an easily identifiable purpose and main idea(s).}}

-{\sethlcolor{green!25}\hl{predictable or overly-obvious main ideas}}; or points that echo observations heard
elsewhere; or a close retelling of another work.

-{\sethlcolor{green!25}\hl{support that is attempted, but developmental details are often limited, uneven,
somewhat off-topic}}, predictable, or too general (e.g., a list of underdeveloped points).\\ \bottomrule
\end{tabular}%
}
\caption{Case study illustrating an essay, high- and low-quality feedback for the Sentence Fluency, Word Choice, Conventions, Organization, and Content traits filtered by FeedEval, and the rubric descriptions corresponding to human-annotated scores. Due to space constraints, we present only the portions of the rubric descriptions that align with the feedback.}
\label{table_case_feedback}
\end{table*}

\subsection{Case Study of Revised Essays} \label{appendix_case_study_esssay}

Table \ref{table_case_revision} presents revised essays produced by Qwen2-1.5B-Instruct based on feedback of different quality for the same original human-written essay, along with the corresponding scores assigned by an automated scoring model (Qwen3-8B). In the essays, revisions corresponding to each trait are highlighted using the same color assigned to that trait. For the {\sethlcolor{cyan!15}\hl{Organization}} trait, no substantial  differences are observed between high- and low-quality feedback, and the resulting revised essays receive identical scores for this trait.

In contrast, for the remaining four traits, differences in feedback quality are clearly reflected in the revised essays and are further manifested in score differences.
First, for the {\sethlcolor{red!20}\hl{Sentence Fluency}} trait, both high- and low-quality feedback refer to the same portions of the original essay when explaining the issues, and the opening sections of the revised essays produced under both feedback conditions are very similar. However, although both feedback types identify areas for improvement, the high-quality feedback offers more concrete and actionable guidance than the low-quality feedback. This difference is reflected in the revisions: while not all revised segments can be explicitly highlighted, the essay revised using high-quality feedback exhibits clearer improvements in the Sentence Fluency trait, including more varied sentence structures, and consequently receives a higher score for the corresponding trait. In contrast, the essay revised using low-quality feedback contains repeated expressions that convey meanings similar to those in preceding sentences ($\textcolor{blue}{\blacksquare}$).

For the {\sethlcolor{yellow!40}\hl{Word Choice}} trait, issues explicitly identified in each feedback are appropriately revised in the corresponding essays. However, aspects mentioned in the high-quality feedback but omitted in the low-quality feedback remain unaddressed in the revised essay guided by low-quality feedback, as shown in ($\textcolor{red}{\star}$). 

Next, for the {\sethlcolor{green!20}\hl{Conventions}} trait, the high-quality feedback explicitly references specific contents of the essay, whereas the low-quality feedback lacks such specificity. As a result, the essay revised using high-quality feedback exhibits more varied sentence structures by rephrasing parts of the initial essay that contained grammatical errors and punctuation mistakes. In contrast, although revisions based on low-quality feedback reduce some conventions-related errors, they primarily involve deleting problematic sentences without introducing newly reformulated content, resulting in relatively limited sentence variation. Not all revised segments can be explicitly highlighted.

Finally, for the {\sethlcolor{blue!15}\hl{Content}} trait, the high-quality feedback provides more concrete and actionable guidance on how to improve the essay than the low-quality feedback. Consequently, the essay revised with high-quality feedback presents more explicit and well-supported evidence to substantiate the author’s main arguments.

\begin{table*}[]
\centering
\scriptsize
\resizebox{\textwidth}{!}{%
\begin{tabular}{|c|m{0.85\textwidth}|}
\toprule
\makecell[c]{\textbf{Initial} \\ \textbf{Trait Scores}}                & \multicolumn{1}{c|}{Sentence Fluency (Score: 1/6), Word Choice (Score: 1/6), Conventions (Score: 1/6), Organization (Score: 1/6), Content (Score: 1/6)} \\ \toprule
\textbf{Topic Prompt}                 & More and more people use computers, but not everyone agrees that this benefits society. Those who support advances in technology believe that computers have a positive effect on people. They teach hand-eye coordination, give people the ability to learn about faraway places and people, and even allow people to talk online with other people. Others have different ideas. Some experts are concerned that people are spending too much time on their computers and less time exercising, enjoying nature, and interacting with family and friends. Write a letter to your local newspaper in which you state your opinion on the effects computers have on people. Persuade the readers to agree with you.                                               \\ \midrule
\makecell[c]{\textbf{Initial} \\ \textbf{Human-written Essay}}                 & More and more people use computers, they love computers for place that are far or the news. Computers are good for everyone in this world. Know one can live with out computers. There's others dat don't like computer for nothing. I know that there's family or friends that be having computers they write each others like in myspace u could have friend from everywhere or family's. But computers are spending to many times on their computers.                                               \\ \midrule
\makecell[c]{\textbf{High-quality} \\ \textbf{Feedback}} &   \textbf{Sentence Fluency}: The essay lacks fluidity and coherence, with sentences such as {\sethlcolor{red!20}\hl{`Computers are good for everyone in this world. Know one can live with out computers.'}} that are choppy and disconnected. The structure is awkward, making it difficult to follow the argument. {\sethlcolor{red!20}\hl{To improve, the writer should work on creating smooth transitions between ideas and constructing more complex and varied sentences}}.

\textbf{Word Choice}: The essay's word choice is limited and imprecise, with phrases like {\sethlcolor{yellow!40}\hl{`Computers are good for everyone'}} lacking depth and specificity. Misused words such as {\sethlcolor{yellow!40}\hl{`Know one'}} instead of `No one' further obscure clarity.{\sethlcolor{yellow!40}\hl{ Improvement can be achieved by using precise vocabulary and ensuring correct word usage to convey ideas more effectively}}.

\textbf{Conventions}: Numerous grammatical errors such as {\sethlcolor{green!20}\hl{`Know one can live with out computers.'}} and {\sethlcolor{green!20}\hl{`There's others dat don't like computer'}} impair readability. The frequent spelling errors ({\sethlcolor{green!20}\hl{`dat'}}) and punctuation issues indicate a lack of control over conventions. {\sethlcolor{green!20}\hl{Thorough proofreading and editing to correct these errors would improve readability.}}

\textbf{Organization}: {\sethlcolor{cyan!15}\hl{The essay lacks a clear organizational structure}}, evidenced by the mix of different ideas in sentences like `There's others dat don't like computer for nothing.' The lack of clear paragraphs and transitions leads to a disjointed reading experience. {\sethlcolor{cyan!15}\hl{Establishing a clear introduction, body, and conclusion and organizing related ideas into paragraphs would improve coherence}}.

\textbf{Content}: The content is underdeveloped with vague ideas, lacking depth and specificity in addressing the essay prompt. Statements like {\sethlcolor{blue!15}\hl{`Computers are good for everyone in this world'}} are too broad and unsupported by examples or discussion. {\sethlcolor{blue!15}\hl{To enhance content, focus on presenting a clear opinion with well-developed arguments supported by concrete details and examples}}.\\ \midrule

\makecell[c]{\textbf{Revised Essay} \\ \textbf{with} \\ \textbf{High-quality} \\ \textbf{Feedback}}  &   {\sethlcolor{red!20}\hl{More and more people use computers, but not everyone agrees that this benefits society. Those who support advances in technology believe that computers have a positive effect on people. They teach hand-eye coordination, give people the ability to learn about faraway places and people, and even allow people to talk online with other people.}} {\sethlcolor{yellow!40}\hl{Others have different ideas. Some experts are concerned that people are spending too much time on their computers and less time exercising, enjoying nature, and interacting with family and friends.}} However, I strongly disagree with this viewpoint. I firmly believe that computers can be a powerful tool that has a profound impact on people's lives. {\sethlcolor{blue!15}\hl{Computers have numerous benefits. They have made it possible for people to access vast amounts of information, allowing them to learn about various subjects and cultures. The internet has also enabled people to connect with others across the globe, fostering global understanding and collaboration. Furthermore, computers have greatly improved communication skills, enabling people to express themselves more effectively and engage in meaningful conversations. In addition, computers have the power to enhance creativity and productivity. They offer a wide range of tools and software that can be used for creative pursuits, such as writing, art, and music. Moreover, computers have improved access to healthcare, education, and employment opportunities, which have greatly benefited society as a whole.}} I firmly believe that computers are a valuable resource that can bring people together and improve their lives in numerous ways. While there are valid concerns about the negative effects of excessive computer use, I firmly disagree that this outweighs the benefits. By harnessing the power of technology, we can create a brighter, more connected, and more prosperous world for everyone.

\textbf{(Sentence Fluency: 4/6, Word Choice: 4/6, Conventions: 4/6, Organization: 4/6, Content: 4/6)}
\\ \midrule

\makecell[c]{\textbf{Low-quality} \\ \textbf{Feedback}}  &   \textbf{Sentence Fluency}: The essay is difficult to follow, with awkward constructions such as {\sethlcolor{red!20}\hl{`Computers are spending to many times on their computers.'}} The sentences lack flow and structure, making the text jarring and irregular. To improve, {\sethlcolor{red!20}\hl{the writer should focus on enhancing the fluidity of sentence structures}}.

\textbf{Word Choice}: The essay employs vague and imprecise language, lacking specificity and reducing clarity for the readers. The limited vocabulary limits clarity and weakens the impact of the essay. {\sethlcolor{yellow!40}\hl{For improvement, the writer should use precise and varied vocabulary to better convey the intended message}}.

\textbf{Conventions}: There are numerous grammatical errors and incorrect convention usage, as seen in the inconsistency with punctuation and capitalization. These errors distract from the readability. {\sethlcolor{green!20}\hl{Focus on proofreading to correct the grammatical mistakes and incorrect punctuation to improve clarity.}}

\textbf{Organization}: {\sethlcolor{cyan!15}\hl{The essay lacks a clear organizational structure}}, with ideas not clearly separated or presented in a logical sequence. It is written as a single fragmented paragraph without identifying distinct ideas or paragraph boundaries. {\sethlcolor{cyan!15}\hl{The absence of clear introduction, body, and conclusion and overall coherence hinders the reader's ability to follow the argument}}. 

\textbf{Content}: The essay presents a limited and underdeveloped exploration of the topic, with vague and general statements such as {\sethlcolor{blue!15}\hl{`Computers are good for everyone in this world.'}} The main ideas are not clearly articulated and underdeveloped, limiting the essay’s persuasiveness.
\\ \midrule

\makecell[c]{\textbf{Revised Essay} \\ \textbf{with} \\ \textbf{Low-quality} \\ \textbf{Feedback}}    & {\sethlcolor{red!20}\hl{More and more people use computers,}} \underline{($\textcolor{red}{\star}$) and computers are good for everyone in this world.} {\sethlcolor{red!20}\hl{Those who support advances in technology believe that computers have a positive effect on people. They teach hand-eye coordination, give people the ability to learn about faraway places and people, and allow people to talk online with other people.}} {\sethlcolor{yellow!40}\hl{However, others have different ideas. Some experts are concerned that people are spending too much time on their computers and less time exercising, enjoying nature, and interacting with family and friends.}} \underline{($\textcolor{blue}{\blacksquare}$) The widespread use of computers has led to a loss of face-to-face interactions and social skills}, which are essential for building strong relationships and understanding each other. While computers can be a valuable tool for learning and communication, they can also have negative consequences. The constant exposure to screens and the lack of physical activity can lead to a sedentary lifestyle, which can have serious health risks. Furthermore, excessive computer use can cause social isolation and decreased social skills, as people spend more time interacting with computers than with others. {\sethlcolor{blue!15}\hl{Despite these concerns, the benefits of technology extend beyond just the individual. Computers have the power to bring people together and facilitate global communication, allowing people to connect with others across the world. They have also enabled people to access a vast amount of information and knowledge, which can be a valuable resource for learning and personal growth.}}

\textbf{(Sentence Fluency: 3/6, Word Choice: 3/6, Conventions: 3/6, Organization: 4/6, Content: 3/6)}
\\ \bottomrule
\end{tabular}%
}
\caption{Case study illustrating essays revised using high- and low-quality feedback filtered by FeedEval across the Sentence Fluency, Word Choice, Conventions, Organization, and Content traits, along with scores assigned by an automated essay scoring model.}
\label{table_case_revision}
\end{table*}

\subsection{Essay Improvement after Revision using Feedback Provided by Trained Models}

To evaluate the pedagogical effectiveness of essay feedback generated by models trained on high- and low-quality feedback filtered by FeedEval and GPT-5.1, we conduct essay revision experiments using small-sized LLMs (Llama3-1B-Instruct and Qwen2-1.5B-Instruct) together with a fine-tuned essay scoring model (Qwen3-8B). All experimental settings are identical to those in section \ref{llm_essay_revision}, except for the source of feedback used for revision.

\begin{figure}[htbp]
    \centering
    \includegraphics[width=\linewidth]{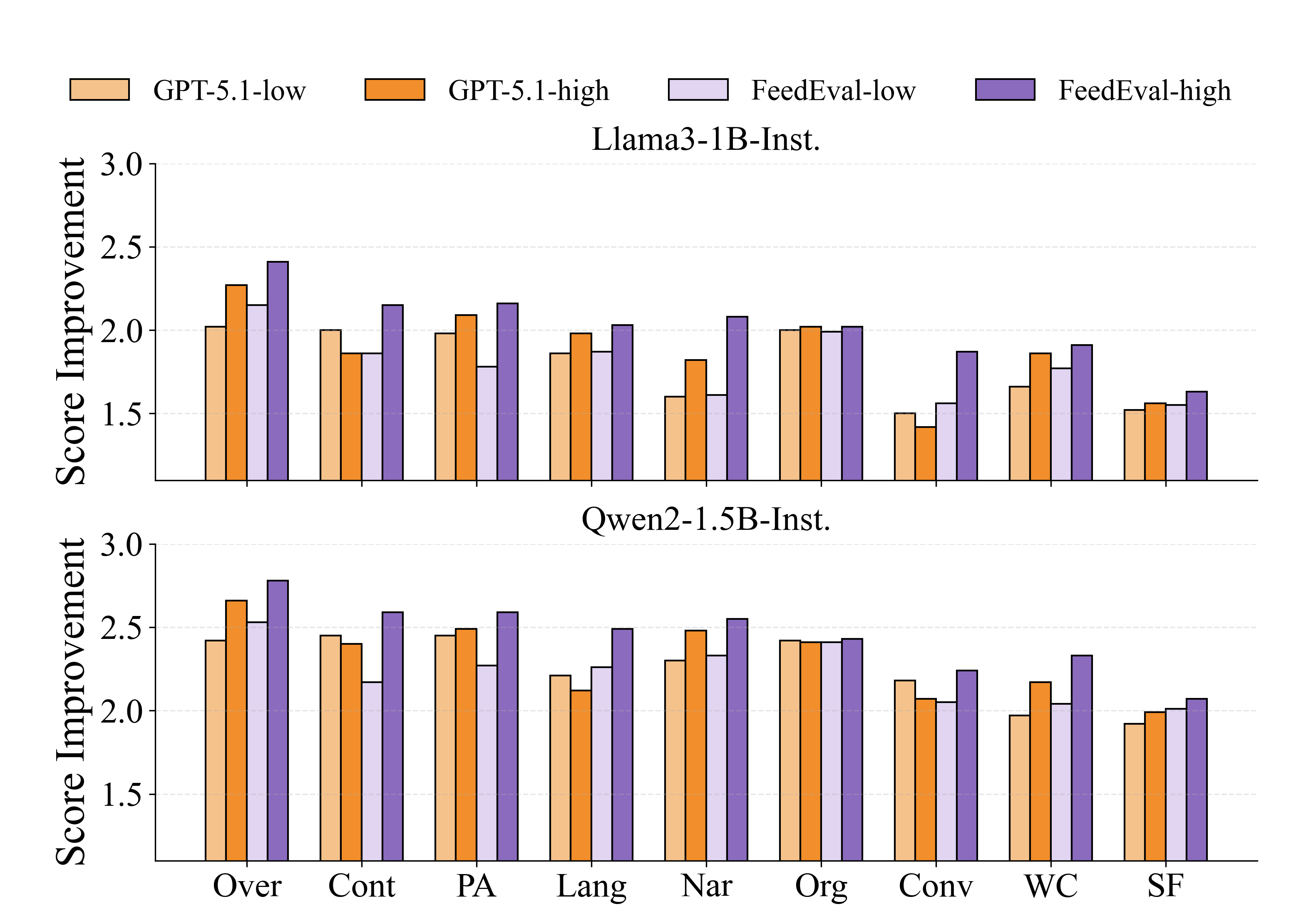}
    \caption{Average essay score improvement across traits on ASAP++ after revisions guided by models trained on feedback labels of high- and low-quality identified by FeedEval and GPT-5.1.}
    \label{fig_trained_revision}
\end{figure}

Specifically, in this experiment, the small LLMs revise essays using feedback generated by Qwen3-8B models trained on LLM-generated feedback filtered by either FeedEval or GPT-5.1. These trained models are the same ones analyzed in section \ref{scoring_performance}, which are trained to jointly generate essay scores and feedback. The feedback is provided in a structured format with trait scores, as described in section \ref{experiment_essay_scoring}. In contrast, in section \ref{llm_essay_revision}, the small LLMs directly receive LLM-generated feedback filtered by FeedEval or GPT-5.1 without involving any intermediate model training. Figure \ref{fig_trained_revision} presents the results.

Overall, revisions guided by feedback generated by the model trained on FeedEval-filtered high-quality feedback exhibit trends consistent with those observed in section \ref{llm_essay_revision}, yielding larger revision gains than feedback guided by the same model trained on low-quality feedback. By comparison, feedback provided by the model trained on GPT-5.1-selected high-quality feedback does not consistently yield larger revision gains than its low-quality counterpart. Across traits, feedback produced by the model trained on FeedEval-selected high-quality feedback results in greater revision gains than feedback generated by the model trained on GPT-5.1-selected high- or low-quality data.

In summary, consistent with the findings from essay scoring experiments, when FeedEval-selected high-quality feedback is used as a structured supervision signal together with scores, it serves as an effective training signal for essay feedback generation and leads to more meaningful essay revisions.

\section{Additional Materials} \label{addtional_materials}

\subsection{Dataset Statistic}
We present a comprehensive statistical analysis of the datasets curated to train dimension-specific evaluator LLMs in Table \ref{table_dimension_data}. We further report statistics of FeedEval-identified high- and low-quality essay feedback on the ASAP++ dataset in Table \ref{table_quality_data}.

\begin{table*}[]
\centering
\small
\begin{tabular}{|c|c|c|c|c|}
\toprule
\textbf{Dataset} & \textbf{Task-type} & \textbf{\begin{tabular}[c]{@{}c@{}}\# of\\Feedback Pair\end{tabular}} & \textbf{\begin{tabular}[c]{@{}c@{}}Chosen Feedback\\Word Count\\(Min-Max)\end{tabular}} & \textbf{\begin{tabular}[c]{@{}c@{}}Rejected Feedback\\Word Count\\(Min-Max)\end{tabular}} \\ \midrule
Specificity (SpecEval) & Rewarding & 41730 & 51.93 (16 - 129) & 49.76 (13 - 129) \\
Helpfulness            & Rewarding & 14158 & 160.08 (7 - 741) & 100.77 (1 - 968) \\
Validity               & NLI       & 99952 & 104.58 (1 - 273) & 104.58 (1 - 273) \\ \bottomrule
\end{tabular}%
\caption{Statistics of the dimension-specific training datasets, including the training task type, number of feedback pairs, and average word counts of chosen and rejected feedback. For the validity dataset, the chosen and rejected feedback correspond to feedback labeled as entailment and contradiction, respectively.}
\label{table_dimension_data}
\end{table*}

\begin{table*}[]
\centering
\resizebox{\textwidth}{!}{%
\begin{tabular}{|c|c|c|c|c|c|}
\toprule
\textbf{Dataset} &
  \textbf{\begin{tabular}[c]{@{}c@{}}Word Count\\ (Min-Max)\end{tabular}} &
  \textbf{\begin{tabular}[c]{@{}c@{}}Specificity Score\\ (Min-Max)\end{tabular}} &
  \textbf{\begin{tabular}[c]{@{}c@{}}Helpfulness Score\\ (Min-Max)\end{tabular}} &
  \textbf{\begin{tabular}[c]{@{}c@{}}Validity Score\\ (Min-Max)\end{tabular}} &
  \textbf{\begin{tabular}[c]{@{}c@{}}Avg. Score\\ (Min-Max)\end{tabular}} \\ \midrule
High-quality &
  256.09 (115 - 455) &
  0.574 (0.0 - 1.000) &
  0.423 (0.0 - 0.999) &
  0.202 (0.084 - 0.405) &
  0.400 (0.207 - 0.786) \\
Low-quality &
  271.61 (158 - 394) &
  0.029 (0.0 - 0.354) &
  0.048 (0.0 - 0.405) &
  0.189 (0.084 - 0.404) &
  0.089 (0.030 - 0.195) \\ \bottomrule
\end{tabular}%
}
\caption{Statistics of the FeedEval-identified high- and low-quality essay feedback datasets, including average word counts, FeedEval scores for specificity, helpfulness, and validity, as well as the average score across the three dimensions.}
\label{table_quality_data}
\end{table*}

\subsection{Materials for Training Essay Evaluation Models}

Table \ref{table_input_output_scores} and \ref{table_input_output_score_feedback} demonstrates the inputs (user prompt) and outputs (assistant prompt) for training essay evaluation models.

\begin{table*}[]
\centering
\small
\resizebox{\textwidth}{!}{%
\begin{tabular}{|c|m{0.85\textwidth}|}
\toprule
\textbf{Input (Score Only)} &  $<$|begin\_of\_text|$><$|start\_header\_id|$>$system$<$|end\_header\_id|$>$

You are an essay evaluator. You will receive an essay and you will need to evaluate the essay of prompt 1, focusing on the following traits: [`sentence fluency', `word choice', `conventions', `organization', `content']. Score the essay in JSON format, using the trait names as keys, without any additional text.$<$|eot\_id|$><$|start\_header\_id|$>$user$<$|end\_header\_id|$>$

Essay: Dear @CAPS1 @CAPS2, I believe that using computers will benefit us in many ways like talking and becoming friends will others through websites like facebook and mysace. Using computers can help us find coordibates, locations, and able ourselfs to millions of information. Also computers will benefit us by helping with jobs as in planning a house plan and typing a @NUM1 page report for one of our jobs in less than writing it. Now lets go into the wonder world of technology. Using a computer will help us in life by talking or making friends on line. Many people have myspace, facebooks, aim, these all benefit us by having conversations with one another. Many people believe computers are bad but how can you make friends if you can never talk to them? I am very fortunate for having a computer that can help with not only school work but my social life and how I make friends. Computers help us with finding our locations, coordibates and millions of information online. If we didn't go on the internet a lot we wouldn't know how to go onto websites that @MONTH1 help us with locations and coordinates like @LOCATION1. Would you rather use a computer or be in @LOCATION3. When your supposed to be vacationing in @LOCATION2. Million of information is found on the internet. You can as almost every question and a computer will have it. Would you rather easily draw up a house plan on the computers or take @NUM1 hours doing one by hand with ugly erazer marks all over it, you are garrenteed that to find a job with a drawing like that. Also when appling for a job many workers must write very long papers like a @NUM3 word essay on why this job fits you the most, and many people I know don't like writing @NUM3 words non-stopp for hours when it could take them I hav an a computer. That is why computers we needed a lot now adays. I hope this essay has impacted your descion on computers because they are great machines to work with. The other day I showed my mom how to use a computer and she said it was the greatest invention sense sliced bread! Now go out and buy a computer to help you chat online with friends, find locations and millions of information on one click of the button and help your self with getting a job with neat, prepared, printed work that your boss will love.$<$|eot\_id|$><$|start\_header\_id|$>$assistant$<$|end\_header\_id|$>$                                               \\ \midrule
\textbf{Output (Score Only)}  &   \{`sentence fluency': 3.0, `word choice': 4.0, `conventions': 4.0, `organization': 4.0, `content': 4.0, `overall': 9.0\}$<$|eot\_id|$><$|start\_header\_id|$>$assistant$<$|end\_header\_id|$>$                                             \\ \bottomrule
\end{tabular}%
}
\caption{An example of an input and its corresponding output used to train essay evaluation models to generate multi-trait scores only.}
\label{table_input_output_scores}
\end{table*}

\begin{table*}[]
\centering
\small
\resizebox{\textwidth}{!}{%
\begin{tabular}{|c|m{0.85\textwidth}|}
\hline
\textbf{Input (Score + Feedback)}                 & 
$<$|begin\_of\_text|$><$|start\_header\_id|$>$system$<$|end\_header\_id|$>$

You are an essay evaluator.
You will receive an essay and you will need to evaluate the essay of prompt 1, focusing on the following traits: [`sentence fluency', `word choice', `conventions', `organization', `content']. Output only the literal evaluation in JSON format, using the trait names as keys, without any additional text.$<$|eot\_id|$><$|start\_header\_id|$>$user$<$|end\_header\_id|$>$

Essay: Dear @CAPS1 @CAPS2, I believe that using computers will benefit us in many ways like talking and becoming friends will others through websites like facebook and mysace. Using computers can help us find coordibates, locations, and able ourselfs to millions of information. Also computers will benefit us by helping with jobs as in planning a house plan and typing a @NUM1 page report for one of our jobs in less than writing it. Now lets go into the wonder world of technology. Using a computer will help us in life by talking or making friends on line. Many people have myspace, facebooks, aim, these all benefit us by having conversations with one another. Many people believe computers are bad but how can you make friends if you can never talk to them? I am very fortunate for having a computer that can help with not only school work but my social life and how I make friends. Computers help us with finding our locations, coordibates and millions of information online. If we didn't go on the internet a lot we wouldn't know how to go onto websites that @MONTH1 help us with locations and coordinates like @LOCATION1. Would you rather use a computer or be in @LOCATION3. When your supposed to be vacationing in @LOCATION2. Million of information is found on the internet. You can as almost every question and a computer will have it. Would you rather easily draw up a house plan on the computers or take @NUM1 hours doing one by hand with ugly erazer marks all over it, you are garrenteed that to find a job with a drawing like that. Also when appling for a job many workers must write very long papers like a @NUM3 word essay on why this job fits you the most, and many people I know don't like writing @NUM3 words non-stopp for hours when it could take them I hav an a computer. That is why computers we needed a lot now adays. I hope this essay has impacted your descion on computers because they are great machines to work with. The other day I showed my mom how to use a computer and she said it was the greatest invention sense sliced bread! Now go out and buy a computer to help you chat online with friends, find locations and millions of information on one click of the button and help your self with getting a job with neat, prepared, printed work that your boss will love.$<$|eot\_id|$><$|start\_header\_id|$>$assistant$<$|end\_header\_id|$>$
 \\ \toprule
\textbf{Output (Score + Feedback)}                 & 
\{`sentence fluency': \{`score': 3.0, `rationale': ``Quotations such as `I believe that using computers will benefit us in many ways like talking and becoming friends will others through websites like facebook and mysace.' and `Computers help us with finding our locations, coordibates and millions of information online.' demonstrate a lack of fluidity, as grammatical errors and awkward phrasing disrupt the reading. Sentences often lack variation in structure and length, 
contributing to a mechanical feeling rather than a natural flow. To improve, the writer should focus on varying sentence structures and ensuring grammatical accuracy to maintain a smoother rhythm and flow.''\}, `word choice': \{`score': 4.0, `rationale': ``Expressions like `wonder world of technology' and `greatest invention sense sliced bread' attempt a conversational tone but come across as clichéd and vague. The repeated use of phrases like `millions of information' lacks precision. More precise and varied language choices would enhance the essay’s impact, making it more engaging and specific.''\}, `conventions': \{`score': 4.0, `rationale': ``The text contains multiple spelling and grammatical errors such as `coordibates,' `garrenteed,' and `appling,' which detract from readability. Capitalization errors like `facebook' and `mysace' also occur. Careful proofreading for spelling, punctuation, and grammar, along with adherence to capitalization norms, would significantly improve clarity.''\}, `organization': \{`score': 4.0, `rationale': ``The essay follows a basic structure with introductory, body, and concluding elements, as demonstrated in `I hope this essay has impacted your descion on computers.' However, the transitions between ideas can be abrupt, and the overall organization could better support the clarity of the argument. Strengthening organization could involve developing clearer paragraph divisions and ensuring each paragraph supports a singular, coherent idea.''\}, `content': \{`score': 4.0, `rationale': ``The content covers favorable aspects of computer use but lacks depth, as seen in the repetitive mention of `talking and becoming friends' without substantial exploration of other potential benefits or drawbacks. Main ideas are present but require further development and detail to engage the reader fully. Enhancing content could involve a more balanced exploration of topics with specific examples and evidence to support claims, addressing different viewpoints for a more rounded perspective.''\}, `overall': \{`score': 9, `rationale': ``nan''\}\}$<$|eot\_id|$><$|start\_header\_id|$>$assistant$<$|end\_header\_id|$>$
\\ \bottomrule
\end{tabular}%
}
\caption{An example of an input and its corresponding output used to train essay evaluation models to jointly generate multi-trait scores and corresponding feedback.}
\label{table_input_output_score_feedback}
\end{table*}

\subsection{Feedback Rating Dimensions used in Human Evaluation}
In section \ref{human_evaluation_feedback_revision}, human experts rated the high- and low-quality essay feedback filtered by FeedEval using the following three dimensions in a 5-likert scale. The dimensions are adopted from \citet{steiss2024comparing}.

\begin{itemize}
    \item \textbf{Faithfulness to essay (D1)}: Does the feedback adequately and accurately reflect the content of the essay?
    \item \textbf{Usefulness for revision (D2)}: Does the feedback sufficiently address areas for improvement?
    \item \textbf{Rubric alignment (D3)}: Is the feedback grounded in the rubric?
\end{itemize}

\end{document}